%% file: VLM-AR3L_arXiv.tex
\documentclass{article}
\usepackage{ijcai26}

\usepackage{times}
\usepackage{soul}
\usepackage{url}
\usepackage[hidelinks]{hyperref}
\usepackage[utf8]{inputenc}
\usepackage[small]{caption}
\usepackage{graphicx}
\usepackage{amsmath}
\usepackage{amsthm}

\usepackage{booktabs}
\usepackage{algorithm}
\usepackage{algorithmic}
\usepackage[switch]{lineno}
\usepackage{amsfonts}
\usepackage{dsfont}
\usepackage{multirow}

\usepackage{color}
\usepackage{xcolor}

\usepackage{placeins}
\usepackage{subcaption}

\title{VLM-AR3L: Vision-Language Models for Absolute and Relative Rewards in Reinforcement Learning}

\author{
Kuan-Chen Chen$^1$\and
Winston Chen$^1$\and
Wei-Fang Sun$^{2}$\And
Min-Chun Hu$^1$\\
\affiliations
$^1$Department of Computer Science, National Tsing Hua University\\
$^2$NVIDIA AI Technology Center (NVAITC)\\
}

\def\Figref#1{Fig.~\ref{#1}}
\def\Tabref#1{Table~\ref{#1}}
\def\Secref#1{Section~\ref{#1}}
\def\Eqref#1{Eq.~(\ref{#1})}

\begin{document}

\maketitle

\begin{abstract}
    Designing effective reward functions remains a major challenge in reinforcement learning (RL), particularly in open-ended environments where task goals are abstract and difficult to quantify. In this work, we present VLM-AR3L, a framework that leverages Vision-Language Models (VLMs) to provide both absolute and relative rewards for RL. VLM-AR3L interprets an agent’s visual observations in the context of a natural language task goal, and learns both absolute and relative rewards from VLM-generated preference labels. The absolute reward model predicts scalar evaluations for individual states, while the relative reward model compares consecutive observations to infer progress or regression toward the task goal. Their integration combines the stability of state-based evaluation with the robustness of comparative supervision. We evaluate VLM-AR3L across benchmarks spanning classic control, manipulation, and open-world embodied tasks, with a particular focus on Minecraft given its visual complexity and long-horizon decision-making requirements. Experimental results show that VLM-AR3L consistently outperforms prior VLM-based reward learning methods. Videos and code are available on the project website: \url{https://vlm-ar3l.github.io/}.
\end{abstract}

\section{Introduction}

Designing effective reward functions remains a major challenge in reinforcement learning (RL)~\cite{laud2004rewardshaping,leike2018scalableagentalignmentreward,openai2019dota2largescale,NEURIPS2022_6255f223}. This challenge is especially pronounced in open-ended or visually complex environments, where task goals are often abstract and difficult to specify precisely. For example, in open-world environments such as Minecraft, agents are often assigned high-level goals (e.g., “build a shelter”) that require long-horizon planning and semantic understanding, while intermediate states provide little explicit reward signal.
As a result, hand-crafting reward functions in such settings is typically impractical, due to ambiguous objectives, sparse feedback, and the absence of privileged state information. To address this bottleneck, recent approaches have explored leveraging large pre-trained models as sources of reward, enabling agents to learn from high-level task descriptions or multimodal alignment signals.

Large language models (LLMs) have been shown to provide structured code or textual feedback for RL agents in text-based or programmatic settings. However, these methods often rely on privileged state access or handcrafted programmatic interfaces, limiting their applicability in real-world visual domains.
When working with visual observations, contrastive vision-language models (cVLMs) such as CLIP are widely used to compute similarity scores between observations and goals for reward shaping. While simple and scalable, these methods typically require task-specific fine-tuning or retraining to be effective across different domains.

\begin{figure}[t]
  \centering
  \includegraphics[width=1\linewidth]{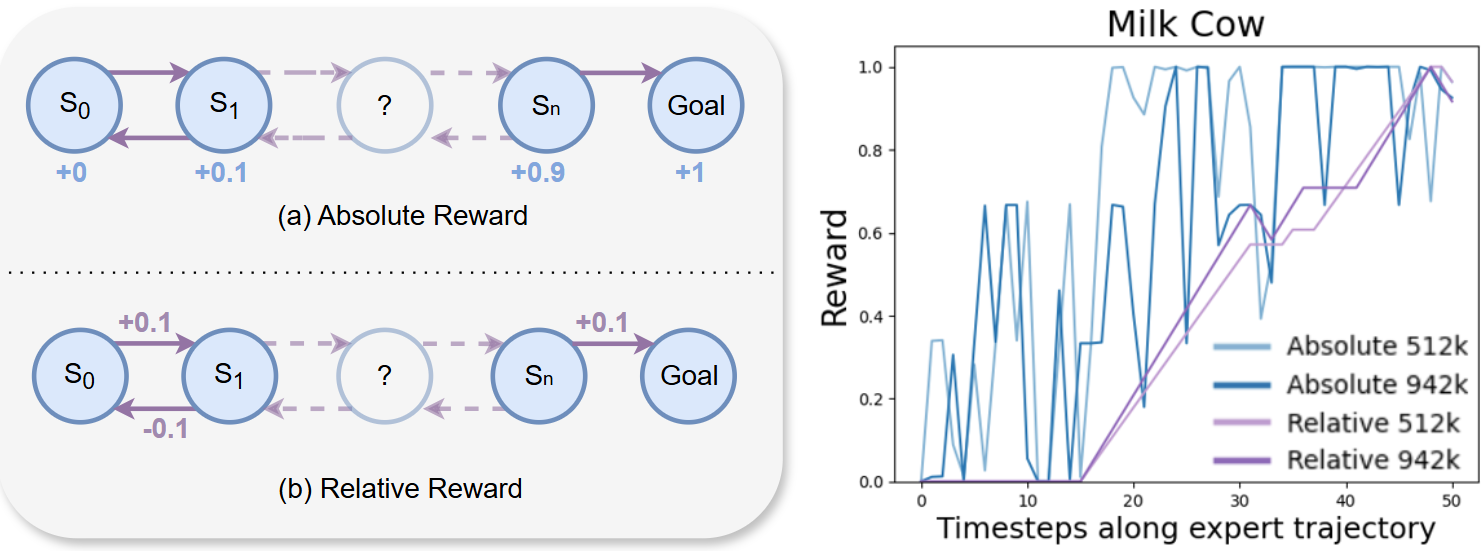}
  \caption{
  \textbf{Comparison of absolute and relative reward formulations.}
  \textbf{Left:} Conceptual illustration in a simple Markov process.
  Absolute reward assigns scalar values to states, where higher scores indicate proximity to the goal.
  Relative reward supervises transitions by assessing progress between state pairs.
  \textbf{Right:} Empirical reward trajectories evaluated along the same expert demonstration at different training steps (512k and 942k), showing that absolute reward varies substantially across training, whereas relative reward, after cumulative summation, exhibits greater temporal consistency.
  }
  \label{fig:relative_reward}
\end{figure}

\begin{figure}[t]
  \centering
  \includegraphics[width=1\linewidth]{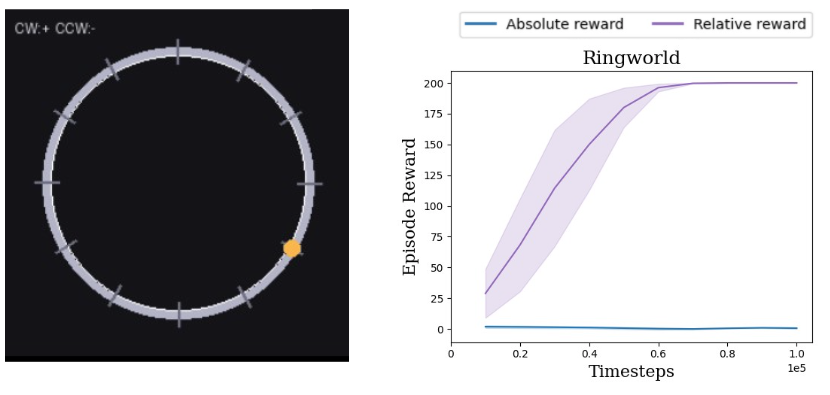}
  \caption{\textbf{RingWorld environment and reward comparison.}
  \textbf{Left:} Illustration of the RingWorld task, where the agent is required to move clockwise along a closed loop.
  \textbf{Right:} Learning curves for agents trained with absolute or relative reward, showing that absolute reward fails under the cyclic state ordering, whereas relative reward enables successful learning.
  }
  \label{fig:RingWorld}
\end{figure}

\begin{figure*}[t]
  \centering
  \includegraphics[width=1\linewidth]{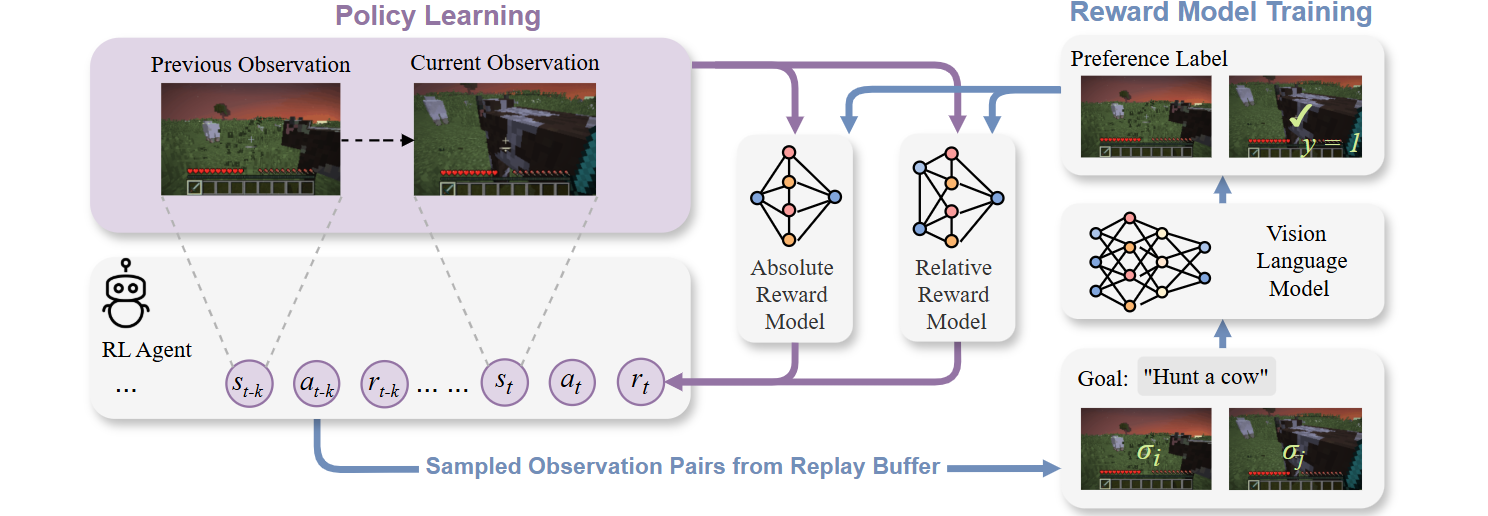}
  \caption{\textbf{Overview of the VLM-AR3L framework.} During reward model training (blue path), observation pairs are sampled from the agent's replay buffer and evaluated by a VLM to determine which observation better aligns with the task goal. The resulting preference labels are used to supervise both absolute and relative reward models. During policy learning (purple path), the trained absolute model evaluates individual observations, while the relative model compares the agent’s current and previous observations to provide complementary absolute and relative reward signals.}
  \label{fig:VLM-AR3L}
\end{figure*}

To move beyond direct similarity-based alignment, preference-based reward modeling has emerged as a promising alternative. These methods learn reward functions by comparing observation pairs and inferring preferences that are typically obtained from human feedback or large pre-trained models. This formulation enables finer-grained feedback and greater robustness in visually diverse or ambiguous environments. For instance, RL-VLM-F \cite{wang2024} leverages generative vision-language models (VLMs) to generate pairwise preferences, which are then used to train a reward model that assigns scalar values to individual states. We call this approach \textbf{absolute reward}. However, we observe that absolute reward signals can be inconsistent across training steps (\Figref{fig:relative_reward}), due to the continual exposure to newly encountered states as training progresses.
These inconsistencies, in turn, make long-horizon tasks difficult to optimize. In particular, absolute reward formulations fundamentally collapse in cyclic task structures or under ambiguous state orderings. In these scenarios, progress cannot be uniquely defined by a global state ordering, a limitation illustrated in \Figref{fig:RingWorld}.
To address these limitations, we propose VLM-AR3L (\Figref{fig:VLM-AR3L}), a framework that leverages VLMs to provide both absolute and relative rewards in reinforcement learning, without requiring fine-tuning or access to privileged internal environment information. Our contributions are as follows:
\begin{itemize}
  \item \textbf{Dual-reward Design:} 
  We introduce a dual-reward framework that jointly learns absolute and relative reward from VLM-generated preferences, with the relative reward providing robustness to reward inconsistency during training (\Figref{fig:relative_reward}) and resolving tasks with ill-defined global state ordering (\Figref{fig:RingWorld}).

 \item \textbf{Relative reward modeling:} 
 We train a Siamese network with VLM-generated preference supervision to model relative reward, avoiding repeated VLM inference during policy optimization, which substantially reduces computational overhead and improves performance.

 \item \textbf{Comprehensive evaluation:} We evaluate VLM-AR3L across diverse domains, including structured robotic control and open-ended tasks. VLM-AR3L consistently outperforms prior VLM-based methods and even surpasses carefully human-designed reward functions. We further conduct detailed ablation studies and VLM accuracy analyses to assess the reliability and scalability of our approach.
\end{itemize}

\section{Related Work}
\paragraph{LLM-based Reward Learning.}
Recent research has explored the use of large pre-trained models, particularly large language models (LLMs), as reward functions for reinforcement learning agents. LLM-based approaches have demonstrated the ability to generate structured code for downstream training \cite{ma2024eurekahumanlevelrewarddesign,xie2024textreward,wang2023voyager,yu2023language,wang2023robogen}, or to provide scalar reward signals from textual task descriptions in language-conditioned or text-based environments \cite{kwon2023reward,pmlr-v202-du23f,chu2023acceleratingreinforcementlearningrobotic}. However, many of these methods assume structured environments or require privileged state access, making them difficult to apply in high-dimensional, real-world visual domains. Additionally, for many open-ended or complex tasks, it is inherently challenging to specify reward logic in code, particularly when task-specific abstractions are involved or when full observability is lacking.

\paragraph{CLIP-based Reward Learning.}
When working with visual observations, a common strategy is to use CLIP-based vision-language models (cVLMs) to compute similarity-based scores between agent observations and language goals, typically via cosine similarity in the shared embedding space \cite{cui2022zeroshot,adeniji2023language,rocamonde2024visionlanguage,sontakke2023roboclip,mahmoudieh2022zeroshot,dipalo2023unifiedagentfoundationmodels,baumli2024visionlanguagemodelssourcerewards,nam2023liftunsupervisedreinforcementlearning}. While conceptually simple, these scores are often noisy and fragile, exhibiting high sensitivity to the phrasing of goals.
To mitigate this, methods such as MineDojo \cite{fan2022minedojo} and CLIP4MC \cite{jiang2024reinforcement} fine-tune CLIP on domain-specific datasets (e.g., Minecraft videos) to generate denser and more accurate reward signals. Other approaches like LIV \cite{ma2023liv} and FuRL \cite{fu2024} propose tailored fine-tuning procedures to enhance reward quality.
While these methods reduce dependence on explicit ground-truth reward annotations, they typically require task-specific fine-tuning to capture subtle variations in visual input. Without fine-tuning, CLIP-based methods often fail to provide nuanced or reliable reward signals, limiting their generalization to new tasks or environments.

\paragraph{Preference-based Reward Learning.}
Beyond similarity-based rewards, recent works explore preference-based reward learning, which uses generative VLMs with multimodal reasoning capabilities, such as Gemini or GPT. A notable direction involves generating preference-based supervision without human labels. For instance, Constitutional AI \cite{bai2022constitutional} and Motif \cite{klissarov2023motif} use LLMs to produce preference labels or intrinsic rewards by comparing model outputs under high-level alignment principles. 
Extending this paradigm to the visual domain, RL-VLM-F \cite{wang2024} proposes learning reward functions by querying vision-language models to compare agent observation pairs and generate preference labels, which are then used to supervise reward learning. 
Several follow-up works further explore the use of VLMs for visual preference modeling and feedback generation in various embodied environments \cite{venkataraman2024realworldofflinereinforcementlearning,ghosh2025preferencevlmleveragingvlms,liu2025vlpvisionlanguagepreferencelearning}. However, these methods largely adopt absolute reward formulations, which are fundamentally limited in cyclic task structures or under ambiguous state orderings, where a consistent global ordering of states is difficult to define. Motivated by this limitation, we explore a dual-reward formulation that augments absolute rewards with relative rewards, aiming to improve robustness and scalability in complex, open-ended environments.

\section{Background}
We consider a standard reinforcement learning (RL) setting modeled as a partially observable Markov decision process (POMDP) \cite{KAELBLING199899}, defined by the tuple \((\mathcal{S}, \mathcal{A}, \mathcal{T}, \mathcal{O}, \mathcal{R}, \gamma)\). Here, \(\mathcal{S}\) denotes the set of environment states, \(\mathcal{A}\) denotes the set of possible actions, \(\mathcal{T}\) denotes the (unknown) state transition dynamics, \(\mathcal{O}\) denotes the observation space, \(\mathcal{R}\) denotes the reward function, and \(\gamma \in [0, 1)\) denotes the discount factor that prioritizes immediate rewards over long-term ones. At each time step \(t\), the agent receives an observation \(s_t \in \mathcal{O}\) (e.g., an RGB image) and selects an action \(a_t \in \mathcal{A}\), which causes the environment state in \(\mathcal{S}\) to evolve according to \(\mathcal{T}\). The objective is to learn a policy \(\pi\) that maximizes the expected cumulative discounted reward over time.

\paragraph{Preference-Based Reinforcement Learning.}

Our work builds on the framework of preference-based reinforcement learning (pbRL), where agents learn reward functions from pairwise comparisons over trajectory segments or behavior snippets \cite{christiano2017deep,ibarz2018reward,2021pebble,lee2021bpref}. In pbRL, a trajectory segment \(\sigma\) typically consists of a sequence of states \(\{s_1, \dots, s_H\}\), where $H$ denotes the sequence length of the segment. In this paper, we consider the special case that \(H = 1\), as multi-image or video-based VLM reasoning remains less reliable in complex visual environments. Given a pair of segments \((\sigma_i, \sigma_j)\), an annotator provides a preference label \(y \in \{0, 1, -1\}\), where \(y = 0\) indicates that \(\sigma_i\) is preferred, \(y = 1\) that \(\sigma_j\) is preferred, and \(y = -1\) that the two are equally preferable or incomparable. 
Given a reward function $r$ over the
states, we follow the standard Bradley-Terry model \cite{bradley1952rank} to compute the preference probability of a pair of segments:
\begin{equation}
\label{eq:BTM}
\mathbb{P}[\sigma_i \succ \sigma_j]
=
\frac{
\exp\!\left(r(s_i) \right)
}{
\sum_{k \in \{i,j\}}
\exp\!\left(r(s_k) \right)
}.
\end{equation}

The reward model is then trained by minimizing a cross-entropy loss over a dataset of annotated preferences:
\begin{align}
\label{CELoss}
\mathcal{L}_\text{reward}(\theta)
= - \mathbb{E}_{(\sigma_i, \sigma_j, y) \sim \mathcal{D}}
& \Big[
 \mathds{1}_{\{y = 0\}} \log \mathbb{P}_\theta[\sigma_i \succ \sigma_j] \nonumber \\
& +\, \mathds{1}_{\{y = 1\}} \log \mathbb{P}_\theta[\sigma_j \succ \sigma_i]
\Big],
\end{align}
where $\theta$ denotes the learnable parameters of the reward model.

In typical pbRL, the reward model and policy are trained iteratively: the reward model is updated using newly collected preference annotations, and the policy is optimized with standard RL algorithms using the learned reward model.

\section{Method}
\label{sec:method}

\Figref{fig:VLM-AR3L} illustrates the overall architecture of the proposed VLM-AR3L framework. The training process consists of two intertwined pipelines: (1) reward model training and (2) policy learning.
In the reward model training phase, observation pairs \((\sigma_i, \sigma_j)\) are sampled from the agent’s replay buffer, along with a high-level language goal \(g\). Each triplet \((\sigma_i, \sigma_j, g)\) is passed to a frozen VLM, which outputs a preference label \(y\) indicating which observation better reflects progress toward the goal. These preference labels are used to supervise the training of both an \textbf{absolute reward model} and a \textbf{relative reward model}. During policy learning, the absolute reward model assigns state-based evaluations to the current observation $s_t$, while the relative reward model estimates progress by comparing the current observation $s_t$ with a previous observation $s_{t-k}$, jointly providing dense and informative reward signals at each timestep. These signals guide the agent’s policy updates using standard reinforcement learning algorithms, eliminating the need for hand-crafted rewards or privileged state information.

\subsection{Absolute Reward Model}

Absolute reward models assign scalar values to individual states, where higher scores reflect states that are closer to the goal. As illustrated in \Figref{fig:relative_reward}, this state-based formulation provides a stable evaluation signal when a global ordering over the state space is well-defined, as each state is mapped to a fixed reward value independent of the agent’s trajectory.

Given a parameterized absolute reward function \(F^{\,\mathrm{abs}}_\psi\) over the
states, the model outputs a scalar reward value for each state following the formulation in ~\Eqref{eq:BTM}:
\begin{equation}
\label{eq:absolute-reward-fn}
F^{\,\mathrm{abs}}_\psi(s_t) = r_\psi(s_t),
\end{equation}
where $r_\psi(s_t) \in \mathbb{R}$ denotes the absolute reward assigned to state $s_t$.
The absolute reward at time step $t$ is then given by:
\begin{equation}
\label{eq:absolute-reward}
R^{\,\mathrm{abs}}_\psi(s_t) = F^{\,\mathrm{abs}}_{\psi}(s_t).
\end{equation}

\subsection{Relative Reward Model}
\label{sec:method_siamese}

In contrast to the absolute reward model, the relative reward model evaluates progress between pairs of states. This design reduces reliance on a global understanding of the task space and improves robustness in partially observable or long-horizon scenarios. When new states are introduced and the reward models are updated, the relative reward of previously seen transitions remains stable. In contrast, absolute reward often suffers from instability, where earlier states may receive drastically different scores due to shifts in global normalization or value scaling (\Figref{fig:relative_reward}). Such robustness is especially beneficial in long-horizon settings, where consistent reward signals are essential for learning.

To implement the relative model, we employ a Siamese neural network \cite{NIPS1993_288cc0ff} 
to learn a relative function \(F^{\,\mathrm{rel}}_\phi\).
Given a pair of states $(s_i, s_j)$, the network outputs the preference probability:
\begin{equation}
    F^{\,\mathrm{rel}}_{\phi}(s_i, s_j)=\mathbb{P}_\phi[\sigma_i \succ \sigma_j].
\end{equation}
The relative model is trained using the same loss defined in \Eqref{CELoss}. 
Architectural and training details
are provided in the supplementary material (Appendix~\ref{appendix:siamese}).

During policy learning, the relative model aims to assess whether a transition indicates progress toward the goal \(g\). To improve the model's sensitivity to meaningful changes, particularly when consecutive observations are highly similar, we introduce a temporal offset \(k \geq 1\) and compare the current observation \(s_t\) with a past observation \(s_{t-k}\).
To ensure the reliability of the model's predictions, we apply a symmetric confidence check. Specifically, a positive reward is assigned only if the model's confidence in preferring \(s_t\) over \(s_{t-k}\) exceeds a threshold \(\tau\), and the reverse prediction also indicates that \(s_{t-k}\) is not preferred over \(s_t\) with confidence exceeding \(\tau\). This bidirectional check ensures both strong and consistent preference. Formally, the relative reward at time step $t$ is computed as:
\begin{equation}
\label{eq:relative-reward}
R^{\,\mathrm{rel}}_\phi(s_t, s_{t-k}) =
\begin{cases}
\;\;1, & \text{if } p^+_t > \tau \ \text{and} \ p^-_t < 1-\tau, \\
-1, & \text{if } p^-_t > \tau \ \text{and} \ p^+_t < 1-\tau, \\
\;\;0, & \text{otherwise},
\end{cases}
\end{equation}
\begin{equation*}
\text{where } \quad
p^+_t = F^{\,\mathrm{rel}}_{\phi}(s_t, s_{t-k}),
\quad
p^-_t = F^{\,\mathrm{rel}}_{\phi}(s_{t-k}, s_t).
\end{equation*}

\subsection{Dual-Reward Design}

Absolute and relative reward models offer complementary perspectives, balancing their individual strengths and weaknesses.
Absolute rewards provide stable, state-based evaluations when a global ordering of the state space is well-defined, whereas relative rewards offer robust, transition-level supervision that remains reliable in cyclic, long-horizon, or partially observable settings.
Motivated by this complementarity, we combine both reward signals to guide policy learning. The dual-reward is computed as:
\begin{equation}
\label{eq:combined-reward}
R^{\,\mathrm{dual}}_{\psi,\phi}(s_t, s_{t-k}) = \alpha R^{\,\mathrm{abs}}_\psi(s_t) + (1-\alpha) R^{\,\mathrm{rel}}_\phi(s_t, s_{t-k}),
\end{equation}
where $\alpha$ is the weighting coefficient for the absolute reward signals, and both absolute and relative rewards are normalized to the same bounded range.
Although the framework learns two reward models, both are trained using a single shared VLM querying pipeline. 
The VLM prompting strategy is described in Appendix~\ref{appendix:VLM_prompt}.

\begin{figure}[t]
  \centering
  \includegraphics[width=1\linewidth]{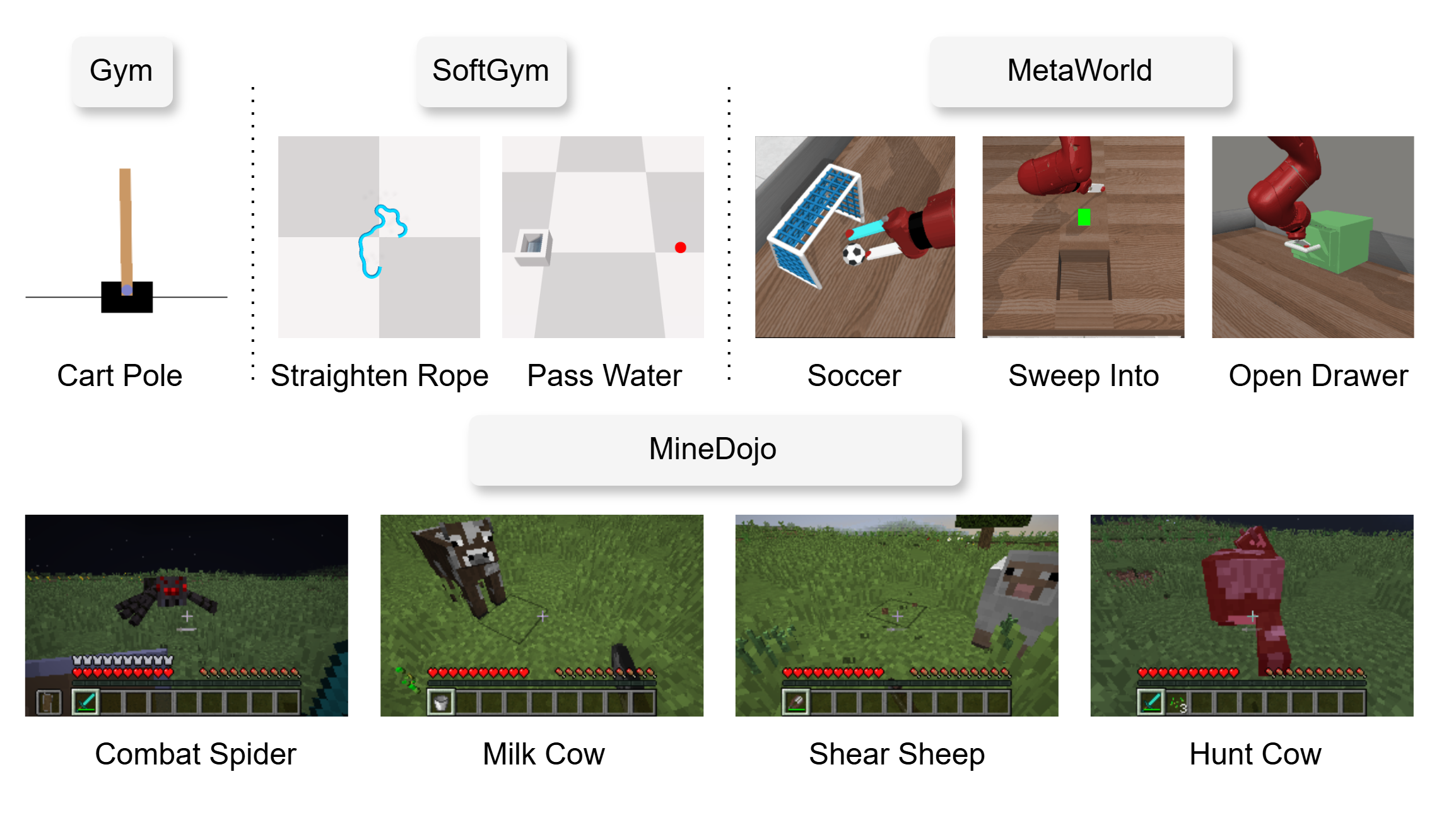}
  \caption{Environments and tasks used in our experiments, spanning structured control (Gym, SoftGym, MetaWorld) and open-ended visual domains (MineDojo).}
  \label{fig:tasks}
\end{figure}

\begin{table*}[t]
\centering
\small
\resizebox{\linewidth}{!}{
\begin{tabular}{l|c|cc|ccc|cccc}
\toprule
\textbf{Model} & \textbf{Cart Pole} & \textbf{Straighten Rope} & \textbf{Pass Water}
& \textbf{Soccer} & \textbf{Sweep Into} & \textbf{Drawer Open} & \textbf{Combat Spider} & \textbf{Milk Cow} & \textbf{Shear Sheep} & \textbf{Hunt Cow} \\
\midrule
Gemini-2.0-Flash \cite{google2024gemini} & \textbf{0.91} & \textbf{0.82} & 0.70 & \textbf{0.87} & \textbf{0.87} & \textbf{0.84} & 0.82 & 0.78 & 0.80 & \textbf{0.76} \\
GPT-4.1-nano \cite{openai2025gpt41nano} & 0.66 & 0.81 & 0.65 & 0.60 & 0.81 & 0.65 & 0.64 & 0.62 & 0.66 & 0.66 \\
Phi-3.5-Vision-Instruct \cite{abdin2024phi3technicalreporthighly}& 0.56 & 0.67 & 0.62 & 0.52 & 0.69 & 0.54 & 0.63 & \textbf{0.90} & \textbf{0.90} & 0.66 \\
MiniCPM-o-2.6 \cite{minicpm2025}& 0.53 & 0.61 & \textbf{0.73} & 0.65 & 0.79 & 0.56 & 0.59 & 0.61 & 0.62 & 0.64 \\
DeepSeek-VL2-Tiny \cite{wu2024deepseekvl2mixtureofexpertsvisionlanguagemodels} & 0.50 & 0.50 & 0.50 & 0.50 & 0.50 & 0.56 & \textbf{0.85} & 0.68 & 0.66 & 0.67 \\
Qwen2.5-VL-7B-Instruct \cite{qwen2.5-VL} & 0.50 & 0.76 & 0.70 & 0.55 & 0.52 & 0.67 & 0.64 & 0.59 & 0.72 & 0.64 \\
InternVL2.5-8B \cite{chen2025expandingperformanceboundariesopensource} & 0.32 & 0.61 & 0.50 & 0.50 & 0.50 & 0.50 & 0.59 & 0.41 & 0.66 & 0.69 \\
Gemma-3-12b-it \cite{gemma_2025} & 0.52 & 0.51 & 0.50 & 0.50 & 0.50 & 0.50 & 0.50 & 0.50 & 0.54 & 0.50 \\
CLIP (ViT-L/14@336px) \cite{pmlr-v139-radford21a} & 0.67 & 0.54 & 0.48 & 0.56 & 0.47 & 0.07 & 0.50 & 0.62 & 0.64 & 0.59 \\
MineCLIP (Attn) \cite{fan2022minedojo}  & --   & --   & --   & --   & --   & --   & 0.45 & 0.56 & 0.55 & 0.39 \\
\bottomrule
\end{tabular}
}
\caption{\textbf{Accuracy of different VLMs in predicting preference labels across tasks.} Each task has 50 positive and 50 negative label queries.}
\label{tab:vlm_accuracy}

\end{table*}

\begin{table*}[t]
\centering
\resizebox{\linewidth}{!}{
\begin{tabular}{l|c|cc|ccc|cccc}
\toprule
\textbf{Methods} & \textbf{Cart Pole} & \textbf{Straighten Rope} & \textbf{Pass Water}
& \textbf{Soccer} & \textbf{Sweep Into} & \textbf{Drawer Open} & \textbf{Combat Spider} & \textbf{Milk Cow} & \textbf{Shear Sheep} & \textbf{Hunt Cow} \\
\midrule
GT Dense (Oracle)   & 1.00 $\pm$ 0.00& 1.00 $\pm$ 0.00& 1.00 $\pm$ 0.00& 1.00 $\pm$ 0.00& 1.00 $\pm$ 0.00& 1.00 $\pm$ 0.00& 0.99 $\pm$ 0.04& 0.38 $\pm$ 0.15& 0.67 $\pm$ 0.25& 0.58 $\pm$ 0.33\\
 
GT Sparse (Oracle)  &  1.00 $\pm$ 0.00& 1.00 $\pm$ 0.00& 1.00 $\pm$ 0.00& 1.00 $\pm$ 0.00& 1.00 $\pm$ 0.00& 1.00 $\pm$ 0.00& 0.84 $\pm$ 0.10& 0.37 $\pm$ 0.01& 0.47 $\pm$ 0.09& 0.15 $\pm$ 0.13\\

CLIP                &  1.00 $\pm$ 0.00& 1.00 $\pm$ 0.00& 0.00 $\pm$ 0.00& 1.00 $\pm$ 0.00& 0.67 $\pm$ 0.47& 0.33 $\pm$ 0.47& 0.59 $\pm$ 0.16& 0.37 $\pm$ 0.00& 0.40 $\pm$ 0.16& 0.02 $\pm$ 0.02\\

MineCLIP & - & - & - & - & - & - & 0.11 $\pm$ 0.09& 0.31 $\pm$ 0.09& 0.20 $\pm$ 0.00& 0.09 $\pm$ 0.08\\

RL-VLM-F            & 1.00 $\pm$ 0.00& 1.00 $\pm$ 0.00& 1.00 $\pm$ 0.00& 1.00 $\pm$ 0.00& 1.00 $\pm$ 0.00& 1.00 $\pm$ 0.00& 0.53 $\pm$ 0.36& 0.50 $\pm$ 0.32& 0.47 $\pm$ 0.38& 0.29 $\pm$ 0.22\\

\midrule

\textbf{VLM-AR3L (Ours)} & \textbf{1.00 $\pm$ 0.00}& \textbf{1.00 $\pm$ 0.00}& \textbf{1.00 $\pm$ 0.00}& \textbf{1.00 $\pm$ 0.00}& \textbf{1.00 $\pm$ 0.00}& \textbf{1.00 $\pm$ 0.00}& \textbf{0.85 $\pm$ 0.05}& \textbf{0.95 $\pm$ 0.00}& \textbf{0.70 $\pm$ 0.10}& \textbf{0.52  $\pm$ 0.02}\\

\bottomrule
\end{tabular}
}
\caption{\textbf{Success rate (mean $\pm$ standard error) of all evaluated methods across tasks.} Results are averaged over 3 random seeds with 5 evaluation episodes per best checkpoint. MineCLIP is only evaluated on MineDojo tasks due to its domain-specific training.}
\label{tab:main_comparison}

\end{table*}

\section{Experiments}
\label{sec:experiments}

\subsection{Setup}

\paragraph{Environments and Tasks.}
To assess the proposed VLM-AR3L, we conduct evaluations on ten tasks across four distinct environments (\Figref{fig:tasks}): \textit{Cart Pole} from Gym \cite{brockman2016openaigym}; \textit{Straighten Rope} and \textit{Pass Water} from SoftGym \cite{corl2020softgym}; \textit{Soccer}, \textit{Sweep Into}, and \textit{Drawer Open} from MetaWorld \cite{yu2019meta}; and \textit{Combat Spider}, \textit{Milk Cow}, \textit{Shear Sheep}, and \textit{Hunt Cow} from MineDojo \cite{fan2022minedojo}.
MineDojo, in particular, features long-horizon, open-ended tasks built on top of Minecraft. These tasks demand multi-step reasoning, complex object interaction, and semantic goal understanding, thereby posing significant challenges for visual reward learning. Collectively, these benchmarks encompass both structured robotic manipulation and expansive open-world scenarios, facilitating a comprehensive evaluation of VLM-AR3L's generalization and adaptability. 
Detailed descriptions of the environments and tasks are provided in the Appendix~\ref{appendix:detail_task}.

\begin{figure*}[t]
  \centering
  \includegraphics[width=1\linewidth]{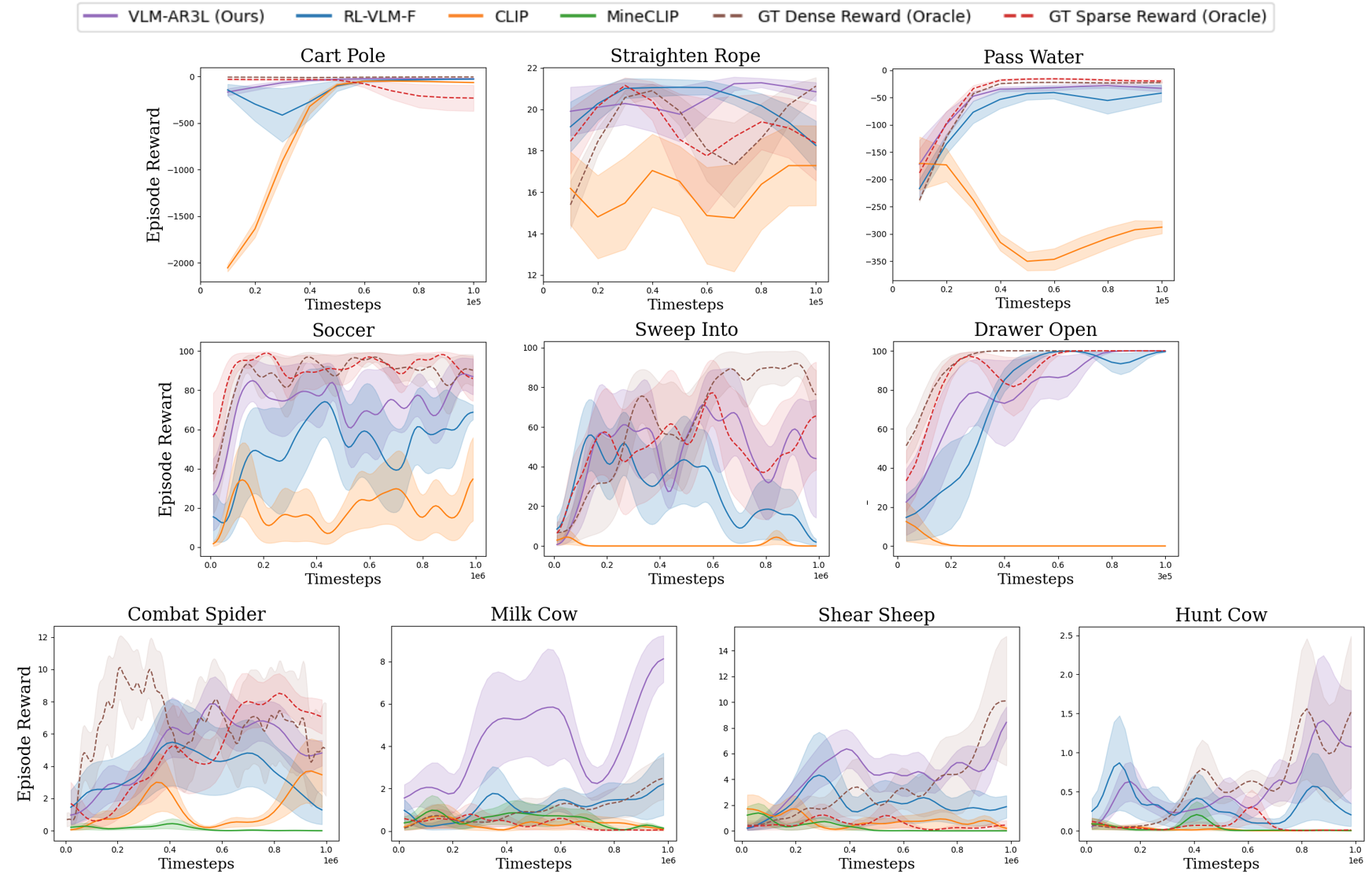}
  \caption{\textbf{Learning curves of all evaluated methods across tasks.} The x-axis represents training timesteps. The y-axis denotes episode rewards  measured by the ground-truth dense reward. Results are averaged over 3 random seeds with 5 evaluation episodes per checkpoint. Shaded regions indicate standard error.}
  \label{fig:compare_result}
\end{figure*}

\paragraph{Baselines.}

We evaluate our approach against several baselines that, like our framework, generate reward functions using only text goals and visual observations, without relying on privileged internal environment state information. In addition, we include oracle baselines using ground-truth rewards to assess whether VLM-generated rewards can serve as practical alternatives to human-designed reward functions.

\begin{itemize}
\item \textbf{RL-VLM-F \cite{wang2024}:} A state-of-the-art preference-based method that queries a VLM for pairwise preferences between observations to train a scalar-valued reward function over individual states.
\item \textbf{CLIP \cite{rocamonde2024visionlanguage}:} A similarity-based method using cosine similarity between CLIP embeddings of the current observation and the language goal for reward shaping.
\item \textbf{MineCLIP \cite{fan2022minedojo}:} A domain-adapted CLIP variant fine-tuned on Minecraft videos, as proposed in MineDojo. Results are not directly comparable to the original paper, as self-imitation learning is omitted to ensure a fair and consistent training protocol across baselines.
\item \textbf{GT Dense Reward (Oracle):} Hand-crafted, incremental reward functions based on full access to environment ground truth.
\item \textbf{GT Sparse Reward (Oracle):} Binary success/failure signals given only at episode completion.
\end{itemize}

We exclude FuRL \cite{fu2024}, RoboCLIP \cite{sontakke2023roboclip} and CLIP4MC \cite{jiang2024reinforcement} from our comparisons for the following reasons: FuRL relies on privileged reward signals, which falls outside our problem setting; RoboCLIP has been reported to perform poorly under comparable assumptions (as reported in RL-VLM-F); and CLIP4MC does not release its fine-tuned CLIP model, precluding an identical reproduction. 
Implementation details for all baselines are provided in Appendix~\ref{appendix:detail_baselines}. 
For policy optimization, we employ Soft Actor-Critic (SAC) \cite{haarnoja2017soft} for Gym, SoftGym, and MetaWorld tasks, and Proximal Policy Optimization (PPO) \cite{schulman2017proximalpolicyoptimizationalgorithms} for MineDojo tasks, selected based on environment characteristics. 
Comprehensive training configurations and hyperparameters are detailed in the Appendix~\ref{appendix:policy}.

\subsection{VLM's Label Accuracy}
\label{sec:vlm_accuracy}

To assess the reliability of using VLMs as supervision, we evaluate several VLMs on multi-image reasoning across our tasks.
For each task, we construct 100 image pairs with balanced pseudo preference labels (a 50/50 split) and prompt each VLM to identify the image that better aligns with the task goal.

\Tabref{tab:vlm_accuracy} summarizes the accuracy of the tested VLMs across tasks. This evaluation serves to identify which models are most suitable as automated annotators for learning relative rewards. Gemini-2.0-Flash stands out for its consistently superior performance, achieving over 70\% accuracy in all tasks, reflecting its robust multimodal reasoning capabilities. In contrast, GPT-4.1-nano underperforms relative to Gemini, showing noticeably lower accuracy across most tasks.
Besides Gemini and GPT models, several open-source VLMs demonstrate competitive performance on specific tasks despite lacking general robustness. For example, Phi-3.5-Vision-Instruct achieves 90\% accuracy on the Milk Cow and Shear Sheep tasks, while DeepSeek-VL2-Tiny excels in the Combat Spider task with an accuracy of 85\%. MiniCPM-o-2.6 performs well in Pass Water, reaching 73\%. On the other hand, CLIP and MineCLIP (evaluated only in MineDojo tasks) exhibit weak performance, underscoring their limitations in modeling multi-step interactions and temporal dependencies. 
Additional details and expanded evaluation results are provided in the Appendix~\ref{appendix:VLM_accuracy}.

\subsection{Comparison with Baselines}

Building on the evaluation in \Secref{sec:vlm_accuracy}, we adopt Gemini-2.0-Flash as the primary VLM for both VLM-AR3L and RL-VLM-F, given its consistently superior annotation performance across tasks.
\Tabref{tab:main_comparison} reports the best success rates of all evaluated methods, and \Figref{fig:compare_result} presents the corresponding learning curves. 
We first examine similarity-based baselines such as CLIP and MineCLIP. While CLIP performs well in the low-dimensional Cart Pole environment, it struggles significantly in more complex environments. Notably, MineCLIP fails to generate reliable reward signals despite being fine-tuned on Minecraft videos.
These results align with prior findings \cite{fu2024,jiang2024reinforcement,wang2024}, suggesting that CLIP-based models often lack the granularity necessary to distinguish subtle visual differences, resulting in noisy and unstable reward signals during training. 

Next, we evaluate RL-VLM-F, a preference-based method that learns scalar rewards over individual states. While RL-VLM-F achieves performance comparable to our method on simpler tasks, its efficacy degrades substantially in long-horizon and visually complex domains such as MineDojo. We attribute this degradation to its reliance on absolute state-based rewards, which requires a global understanding of the task space and often lacks robustness to novel or unseen state variations. In contrast, our proposed VLM-AR3L leverages relative comparisons between transitions, enabling it to generate more stable and goal-consistent rewards even as the state distribution evolves during training.

Finally, when compared against ground-truth dense and sparse rewards, our VLM-AR3L shows strong potential as a practical alternative. In most tasks, VLM-AR3L matches the performance of GT Dense Reward. Moreover, in tasks such as Milk Cow, Shear Sheep, and Hunt Cow, where GT Sparse Reward fails to provide sufficient guidance for learning, VLM-AR3L significantly outperforms the baseline, demonstrating its ability to handle sparse feedback and challenging long-horizon scenarios without requiring handcrafted rewards. These results underscore the practical applicability of our method for real-world settings where reward design is often infeasible.

\begin{figure}[t]
  \centering
  \includegraphics[width=1\linewidth]{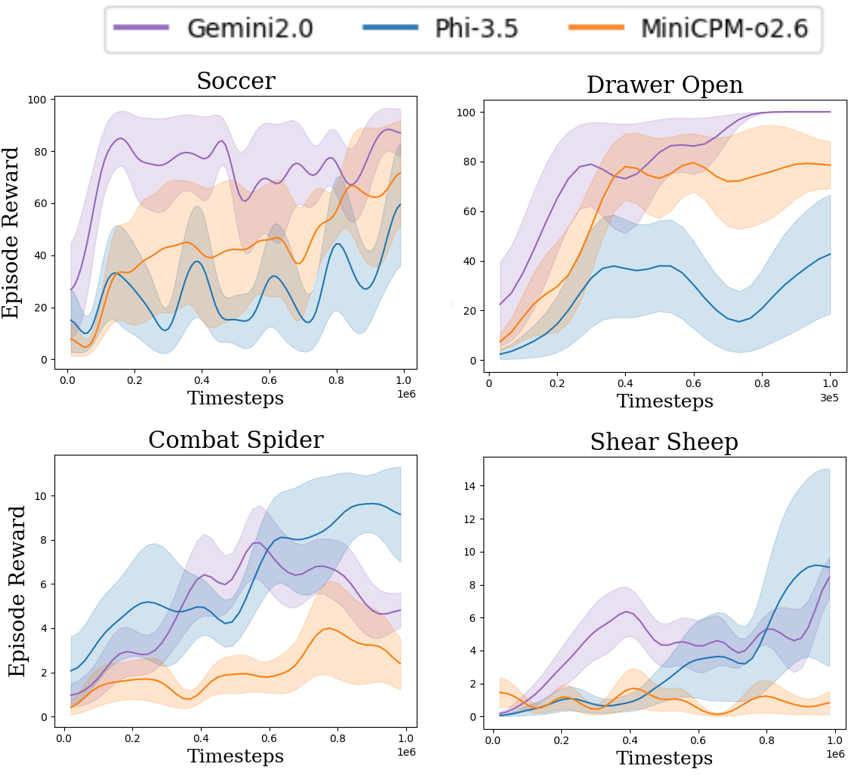}
  \caption{Ablation results with different vision-language models.}
  \label{fig:ablation_VLM_clip}
\end{figure}

\subsection{Ablation Studies}

\paragraph{The Impact of VLM.}
While \Secref{sec:vlm_accuracy} evaluates VLMs as annotators in terms of preference-label accuracy, we further study how the choice of VLM affects end-to-end RL performance when used to generate training supervision for VLM-AR3L.
Specifically, we instantiate the same training pipeline and only vary the annotator VLM (Gemini-2.0, Phi-3.5, MiniCPM-o 2.6) across all tasks (\Figref{fig:ablation_VLM_clip} 
; the full figures and table are provided in the Appendix~\ref{appendix:ablation_vlm}).
Overall, the performance trends are broadly consistent with the label accuracy results in \Secref{sec:vlm_accuracy}.
While Phi-3.5 and MiniCPM-o 2.6 do not excel across all tasks, both exhibit clear task-specific strengths. Given their relatively small model sizes and open-source availability, they offer promising potential for broader adoption in future applications.

\paragraph{The Impact of Relative Reward Architecture.}

\begin{figure}[t]
  \centering
  \includegraphics[width=1\linewidth]{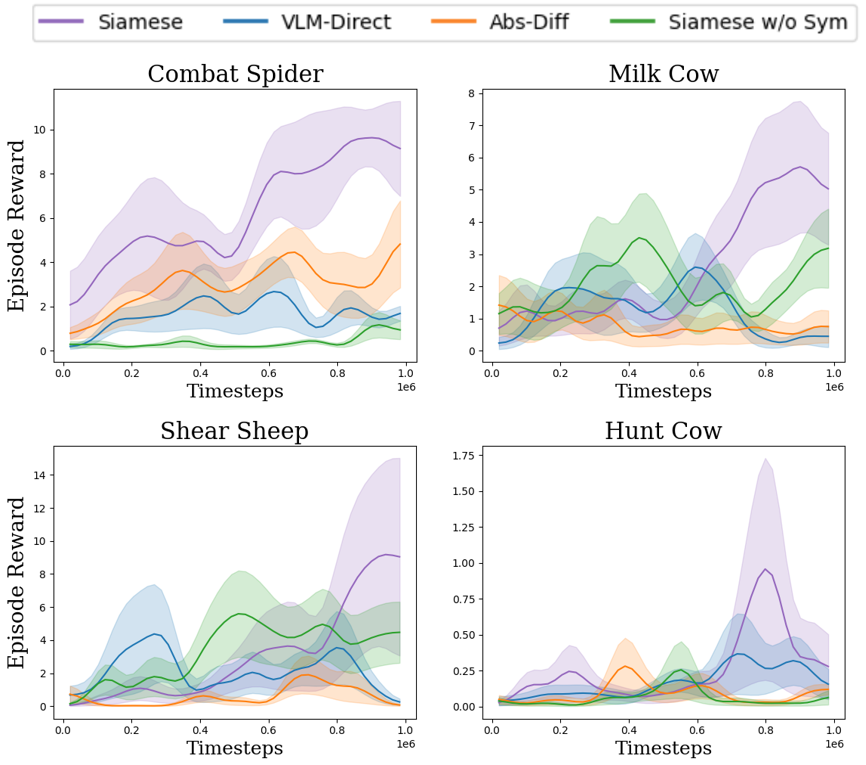}
  \caption{Ablation studies of the relative reward architecture across all experiments, using Phi-3.5 as the VLM.}
  \label{fig:ablation_siamese}
\end{figure}

\begin{figure}[!t]
  \centering
  \includegraphics[width=1\linewidth]{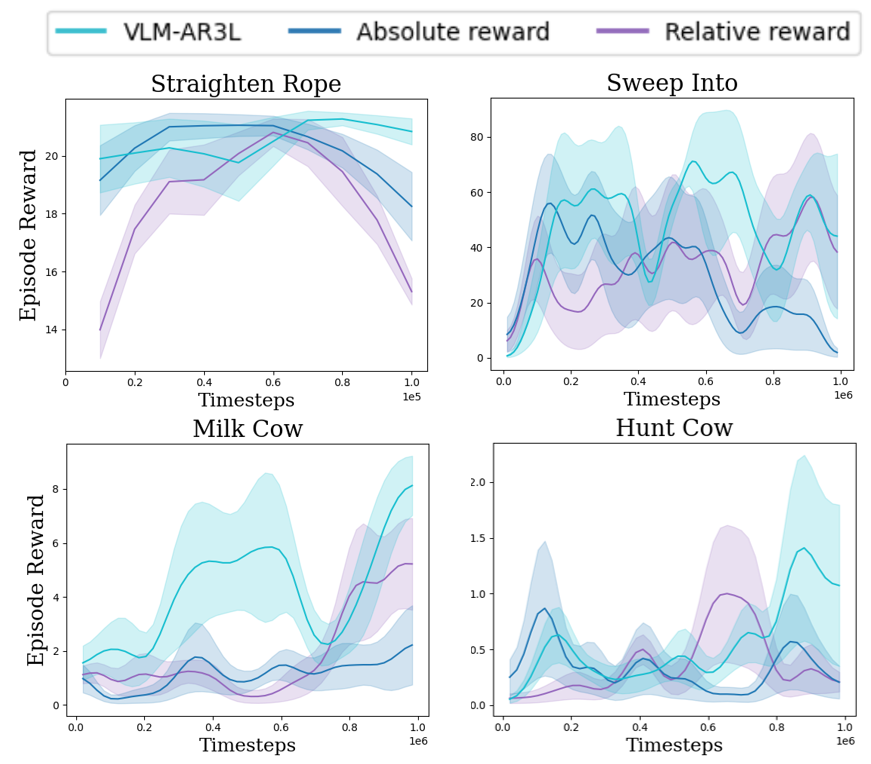}
  \caption{Ablation studies of absolute and relative rewards.}
  \label{fig:ablation_combine_clip}
\end{figure}

An alternative design choice for relative reward, instead of explicitly training a Siamese model (denoted as Siamese in \Figref{fig:ablation_siamese}), is to directly query a VLM at each timestep to evaluate progress (denoted as VLM-Direct).
Another baseline derives relative rewards based on differences between absolute reward predictions (denoted as Abs-Diff).
As shown in \Figref{fig:ablation_siamese}, explicitly training a Siamese model consistently outperforms direct VLM querying during rollouts (VLM-Direct). This improvement stems from preference-based training, which reduces noise in VLM judgments by learning a more stable comparison function, while also lowering computational overhead by avoiding repeated VLM queries at inference time (see Appendix~\ref{appendix:compute}).
We observe that computing relative rewards from absolute reward estimates (Abs-Diff) leads to inferior performance. This is because when the absolute reward model fails to assign reliable scores to individual states, the resulting differences become noisy and unreliable, leading to inaccurate relative assessments.
We also examine the role of the symmetric comparison rule in \Eqref{eq:relative-reward} by evaluating a variant that omits this consistency check (Siamese w/o Sym). Removing the rule leads to notable performance drops across tasks, as it no longer filters inconsistent VLM judgments or prevents biased rewards when observations are visually similar.

\paragraph{The Impact of Absolute and Relative rewards.}

\Figref{fig:ablation_combine_clip} 
(Full figures are provided in the Appendix~\ref{appendix:ablation_combine}) 
analyzes how absolute and relative rewards contribute to each task. Absolute rewards are most beneficial in tasks with a well-defined global progress measure. In contrast, relative rewards become critical in tasks with ambiguous state ordering, cyclic structure, or long-horizon exploration. Combining the two consistently leads to further improvements, indicating that the reward signals are complementary.

\paragraph{The Impact of Hyper-parameters.}

All results above are obtained using a single shared set of hyperparameters, with $\alpha = 0.5$, $\tau = 0.52$, and $k = 16$, without task-specific tuning.
We further confirm this robustness through hyperparameter ablation studies (see Appendix~\ref{appendix:ablation_hyper}), which show only minor performance variations.

\section{Conclusion}
Our results demonstrate that absolute and relative rewards provide complementary supervision for reinforcement learning with VLM-generated preferences. Absolute rewards are effective when a global notion of progress is well defined, while relative rewards are more robust in scenarios involving ambiguous, cyclic, or long-horizon state transitions. Integrating both reward types leads to more stable optimization and consistently outperforms existing VLM-based reward learning methods. Notably, in several tasks, our approach even surpasses manually engineered human reward functions.

\section*{Acknowledgements}
This research was funded by the National Science and Technology Council (NSTC) under grant numbers 114-2218-E-007-019. The last author was partially supported by the Hon Hai Research Institute. We express our sincere appreciation to the NVIDIA AI Technology Center (NVAITC) for the technical support and access to the Taipei-1 supercomputer.
\bibliographystyle{named}
\bibliography{ijcai26}

\include{Appendix}

\end{document}

%% file: Appendix.tex
\appendix
\section{VLM}
\label{appendix:VLM}
\subsection{Prompts}
\label{appendix:VLM_prompt}

To evaluate VLM prompting strategies, we crafted three prompt templates of increasing reasoning complexity:

\begin{itemize}
    \item \textbf{S1 — Single Stage}
          \begin{enumerate}
              \item
                  \texttt{The goal is \{\textit{task\_description}\>\}. 
                  Is the goal better achieved in Image 1 or Image 2? Relay a single line of 1 if the goal is better achieved in Image 1, or 2 if it is better achieved in Image 2. Reply 0 if the text is unsure or there is no difference.}
            \end{enumerate}

    \item \textbf{S2 — Two Stages}
          \begin{enumerate}
              \item
                    \texttt{The goal is \{\textit{task\_description}\>\}. Is the goal better achieved in Image 1 or Image 2?}
              \item 
                    \texttt{Based on the text above, Relay a single line of 1 if the goal is better achieved in Image 1, or 2 if it is better achieved in Image 2. Reply 0 if the text is unsure or there is no difference.}
          \end{enumerate}

    \item \textbf{S2 + CoT — Two Stages with Chain-of-Thought}
          \begin{enumerate}
              \item 
                    \begin{itemize}
                        \item \texttt{What is shown in Image 1?}
                        \item \texttt{What is shown in Image 2?}
                        \item \texttt{The goal is \{<\textit{task\_description}\>\}. Is there any difference between Image 1 and Image 2 in terms of achieving the goal?}
                    \end{itemize}
              \item Same as \emph{S2} Stage 2.
          \end{enumerate}
          
    \item \textbf{S1\textsubscript{2} — Single Stages (Binary)}
          \begin{enumerate}
              \item
                  \texttt{The goal is \{\textit{task\_description}\>\}. 
                  Is Image 2 more likely to achieve the goal? 
                    Reply a single line of 1 if yes, otherwise 0.}
          \end{enumerate}

    \item \textbf{S2\textsubscript{2} — Two Stages (Binary)}
          \begin{enumerate}
              \item
                    \texttt{The goal is \{\textit{task\_description}\>\}.  Is Image 2 more likely to achieve the goal?}
              \item 
                    \texttt{Based on the text above, Reply a single line of 1 if yes, otherwise 0.}
          \end{enumerate}
\end{itemize}

\paragraph{Rationale.}
\begin{itemize}
    \item \textbf{Single-Stage (S1).}  
          The VLM is asked to output only a categorical label.  
          This setting measures the model’s raw decision quality without exposing its reasoning.

    \item \textbf{Two-Stage (S2).}  
          We first solicit an \emph{open-ended answer} that explains \emph{why} the goal is better achieved in one image.  
          The model then self-produces the final label in Stage 2.  
          Recording Stage 1 lets us inspect the model’s internal rationale and assess whether it is aligned with the label it finally assigns.

    \item \textbf{Two-Stage + CoT (S2 + CoT).}  
          Stage 1 is further broken into three fixed sub-questions (\emph{describe Image 1, describe Image 2, compare w.r.t.\ the goal}).  
          This \emph{structured Chain-of-Thought} (CoT) explicitly guides the model to observe, compare, and reason step-by-step before emitting the label in Stage 2, reducing the chance of shortcut reasoning.

    \item \textbf{Binary (S1\textsubscript{2} and S2\textsubscript{2}).}  
          The VLM is required to output a \emph{binary} label (\texttt{0} or \texttt{1}).  
          Empirically, we find that some models perform better when the prompt is kept as simple as possible.   
\end{itemize}

\subsection{VLM accuracy}
\label{appendix:VLM_accuracy}

\paragraph{Evaluation Protocol.}
For each task we curate 100 image pairs with \emph{pseudo} ground-truth preference labels:  
\begin{itemize}
    \item \textbf{50 pairs} in which Image~2 represents an improvement over Image~1 (\texttt{label\;=\,1}), and  
    \item \textbf{50 pairs} in which it does not (\texttt{label\;=\,0}).\footnote{%
          Label 0 subsumes both ``Image 1 is better than Image 2’’ and ``the two images are indistinguishable w.r.t.\ the goal.’’}
\end{itemize}

The VLM’s raw prediction $y\!\in\!\{0,1,2\}$ is post-processed by
\[
\hat{y}=\max(y-1,\,0),
\]
so that  
\begin{itemize}
    \item $y=2$ (\emph{Image 2 better}) $\;\longrightarrow\; \hat{y}=1$, and
    \item $y\in\{0,1\}$ (\emph{unsure/no difference} or \emph{Image 1 better}) $\;\longrightarrow\; \hat{y}=0$.
\end{itemize}
We then compare $\hat{y}$ with the pseudo ground-truth label for accuracy and other downstream metrics.

\begin{table}[t]
\centering
\caption{Impact of prompt design on VLM accuracy across ten tasks (higher is better).}
\label{tab:vlm_prompt_results}
\small
\resizebox{\linewidth}{!}{%
\begin{tabular}{ll|c|cc|ccc|cccc}
\toprule
\textbf{Model} & \textbf{Prompt} &
\textbf{CartPole} &
\textbf{Straighten} &
\textbf{Pass} &
\textbf{Soccer} &
\textbf{Sweep} &
\textbf{Drawer} &
\textbf{Combat} &
\textbf{Milk} &
\textbf{Shear} &
\textbf{Hunt} \\
& &
& \textbf{Rope} &
\textbf{Water} &
& \textbf{Into} &
\textbf{Open} &
\textbf{Spider} &
\textbf{Cow} &
\textbf{Sheep} &
\textbf{Cow} \\
\midrule
\multirow{5}{*}{MiniCPM-o-2.6}
  & S1\textsubscript{2} & 0.50 & 0.64 & 0.50 & 0.65 & 0.79 & 0.56 & 0.59 & 0.77 & 0.57 & 0.64 \\
  & S2\textsubscript{2} & 0.45 & 0.61 & 0.54 & 0.80 & 0.61 & 0.64 & 0.65 & 0.61 & 0.65 & 0.62 \\
  & S1                    & 0.50 & 0.60 & 0.44 & 0.53 & 0.54 & 0.60 & 0.53 & 0.72 & 0.72 & 0.53 \\
  & S2                    & 0.52 & 0.50 & 0.54 & 0.53 & 0.73 & 0.60 & 0.65 & 0.70 & 0.58 & 0.70 \\
  & S2 + CoT              & 0.53 & 0.61 & 0.73 & 0.63 & 0.55 & 0.76 & \textbf{0.82} & 0.76 & 0.62 & 0.72 \\
\midrule
\multirow{5}{*}{Phi-3.5}
  & S1\textsubscript{2} & 0.50 & 0.59 & 0.50 & 0.52 & 0.69 & 0.54 & 0.63 & \textbf{0.90} & \textbf{0.90} & 0.66 \\
  & S2\textsubscript{2} & 0.76 & 0.50 & 0.47 & 0.50 & 0.75 & 0.53 & 0.66 & 0.83 & 0.89 & 0.68 \\
  & S1                    & 0.68 & 0.67 & 0.62 & 0.51 & 0.60 & 0.43 & 0.62 & 0.58 & 0.76 & 0.66 \\
  & S2                    & 0.50 & 0.63 & 0.55 & 0.54 & 0.50 & 0.52 & 0.58 & 0.62 & 0.76 & 0.59 \\
  & S2 + CoT              & 0.56 & 0.61 & 0.56 & 0.45 & 0.46 & 0.51 & 0.59 & 0.66 & 0.72 & 0.64 \\
\midrule
\multirow{3}{*}{GPT-4.1-nano}
  & S1                    & 0.48 & 0.50 & 0.49 & 0.84 & 0.50 & 0.51 & 0.58 & 0.59 & 0.60 & 0.53 \\
  & S2                    & 0.43 & 0.67 & 0.58 & 0.50 & 0.50 & 0.41 & 0.49 & 0.58 & 0.62 & 0.59 \\
  & S2 + CoT              & 0.66 & 0.81 & 0.65 & 0.60 & 0.81 & 0.65 & 0.64 & 0.62 & 0.66 & 0.66 \\
\midrule
\multirow{3}{*}{Gemini-2.0}
  & S1                    & \textbf{0.91} & 0.82 & 0.58 & 0.73 & \textbf{0.87} & \textbf{0.84} & \textbf{0.82} & 0.78 & 0.80 & 0.76 \\
  & S2                    & 0.86 & \textbf{0.87} & 0.59 & 0.61 & \textbf{0.87} & 0.64 & 0.72 & 0.78 & 0.79 & 0.73 \\
  & S2 + CoT              & \textbf{0.91} & 0.84 & \textbf{0.70} & \textbf{0.87} & \textbf{0.87} & \textbf{0.84} & 0.62 & 0.76 & 0.77 & \textbf{0.76} \\
\bottomrule
\end{tabular}
}
\end{table}

\begin{table}[t]
\centering
\small
\caption{Details of the VLMs we benchmarked on an RTX 3090 (float16, batch = 1, prompt S1).}
\label{tab:vlm_detail}
\small
\resizebox{\linewidth}{!}{%
\begin{tabular}{l|ccc}
\toprule
\textbf{Model} & \textbf{Params (B)} & \textbf{VRAM (GB)} & \textbf{Latency (s/query)} \\
\midrule
Phi-3.5-Vision-Instruct & 4.15 & 10 & 0.38 \\
MiniCPM-o-2.6           & 8.67 & 18 & 0.34 \\
DeepSeek-VL2-Tiny       & 3.37 & 16 & 0.20 \\
Qwen2.5-VL-7B-Instruct  & 8.29 & 16 & 0.25 \\
InternVL2.5-8B          & 8.08 & 21 & 1.17 \\
Gemma-3-12b-it          & 12.2 & 22 & 10.36 \\
CLIP (ViT-L/14@336px)   & 0.43 & 1 & 0.03 \\
MineCLIP (Attn)         & 0.16 & 2 & 0.02 \\
\bottomrule
\end{tabular}}
\end{table}

Table~\ref{tab:vlm_prompt_results} summarizes the performance of different prompt designs evaluated across multiple models. 
We observed that for \textbf{Gemini~2.0}, even the simplest prompt design (S1) yielded strong performance. In contrast, \textbf{GPT-4.1-nano} performed better with the slightly more elaborate S2 prompt, and showed further improvements in accuracy with Chain-of-Thought prompting (S2 + CoT). Interestingly, \textbf{Phi-3.5} exhibited the opposite trend in MineDojo-style tasks, favoring simpler prompts—achieving the best results with \textbf{S1\textsubscript{2}}, a binary decision format with a single-stage comparison.

As a result, in all subsequent experiments, we adopt the following prompt configuration per model:
\begin{itemize}
    \item \textbf{Gemini~2.0} and \textbf{MiniCPM} use \textbf{S1},
    \item \textbf{Phi-3.5} uses \textbf{S1\textsubscript{2}}.
\end{itemize}

Table~\ref{tab:vlm_detail} provides detailed statistics on computational resources and runtime used for evaluating each model.

\section{Hyper-parameters and Network
 Architectures}
\subsection{Relative Reward Model (Siamese)}
\label{appendix:siamese}
\subsubsection{Architecture}
The overall architecture of the Siamese model is summarized in Table~\ref{tab:siamese_arch}.
Each image is first processed by a \emph{shared} CNN encoder that produces a $128$-dimensional embedding.  
The two embeddings are concatenated and passed through a fully connected layer with $64$ hidden units, followed by a final linear layer that outputs a two-dimensional logits vector indicating the preference direction (e.g., \textit{regression} vs.\ \textit{improvement}).
Because the Siamese architecture is inherently symmetric, we relabel $\,y=-1$ (``equally preferable'' or ``incomparable'') to $\,y=0$ during training, resulting in a binary target $y\in\{0,1\}$.

\begin{table}[t]
  \centering
  \caption{Siamese reward-model architecture.  A shared CNN maps every
           input image to a 128-D embedding; the two embeddings are concatenated
           and processed by an MLP head that outputs two logits.}
  \label{tab:siamese_arch}
  \small
  \begin{tabular}{cccc}
    \toprule
    \# / Stage & Layer                    & Specification                         & Activation \\ \midrule
    \multicolumn{4}{c}{\textit{Shared encoder (per image)}} \\ \midrule
    1–6        & Custom CNN               & see Table~\ref{tab:cnn_arch} (128-D)  & ReLU, Tanh \\
    \midrule
    \multicolumn{4}{c}{\textit{Comparison head (MLP, takes $256=128{\times}2$-D)}} \\ \midrule
    7          & Linear                   & $256{\to}64$                          & ReLU       \\
    8          & Linear                   & $64{\to}2$                            & —          \\
    \bottomrule
  \end{tabular}
\end{table}

\paragraph{Training Details.}
\begin{itemize}
 \item \textbf{Loss}\,: cross-entropy on the two-class logits
  \item \textbf{Ensemble}\,: three independently initialized Siamese models; their softmax outputs are averaged at inference for greater robustness
  \item \textbf{Sampling strategy}\,: uniform from the replay buffer  
  \item \textbf{Sampling frequency}\,: 50 pairs per 1\,000 environment steps  
  \item \textbf{Buffer capacity}\,: 500\,000 transition pairs  
  \item \textbf{Training frequency}\,: 50 gradient updates per 1\,000 environment steps  
  \item \textbf{Batch size}\,: 32  
\end{itemize}

\subsection{Policy model}
\label{appendix:policy}

\paragraph{MineDojo tasks.}
\begin{itemize}
  \item \textbf{Framework}\,: \texttt{stable-baselines3} — Proximal Policy Optimization (PPO)
  \item \textbf{Network}\,: frozen MineCLIP image encoder $\rightarrow$ MLP(64, 64) 
  \item \textbf{Entropy coefficient}\,: $0.01$  
   \item \textbf{Other parameters}\,: default  
  \item \textbf{Vectorised environment}\,: \texttt{DummyVecEnv} with 4 workers  
  \item \textbf{Observation wrapper}\,: 4-frame stack
\end{itemize}

\paragraph{Other tasks.}
\begin{itemize}
  \item \textbf{Framework}\,: modified \texttt{RL-VLM-F} — Soft Actor–Critic (SAC)
\item \textbf{Network}\,: 3 hidden layers × 256 units, \texttt{tanh} activations (actor \& critic)
  \item \textbf{Learning rates}\,: $3\!\times\!10^{-4}$ for both actor and critic  
  \item \textbf{Batch size}\,: 512  
\item \textbf{Replay buffer}\,: 500\,000 transitions

\end{itemize}

\subsection{Compute Resources}
\label{appendix:compute}
All experiments were conducted on NVIDIA L40 GPUs with 48GB memory. We also verified that all experiments can be reproduced on NVIDIA RTX 3090 GPUs with 24GB memory. Each experiment run with a single seed can typically be completed within 24 hours.
By training a Siamese relative reward model, our scheduling strategy reduces the number of VLM queries during policy learning by approximately $20\times$ compared to direct VLM querying at rollout time.

\section{Details on Tasks and Environments}
\label{appendix:detail_task}

\begin{table}[t]
\centering
\small
\caption{Tasks, simulation suites, and goal prompts used in our experiments.}
\label{tab:tasks}
\resizebox{\linewidth}{!}{%
\begin{tabular}{lll}
\toprule
\textbf{Suite} & \textbf{Task ID} & \textbf{Goal prompt supplied to the VLM}\\
\midrule
\multirow{1}{*}{Gym}%
 & \texttt{CartPole} & balance the brown pole on the black cart to be upright \\
\midrule
\multirow{2}{*}{SoftGym}%
 & \texttt{Straighten Rope} & straighten the blue rope \\
 & \texttt{Pass Water} & move the container that holds water as close to the red circle as possible without spilling \\
\midrule
\multirow{3}{*}{MetaWorld}%
 & \texttt{Soccer} & minimise the distance between the soccer ball and the goal \\
 & \texttt{Sweep Into} & minimise the distance between the green cube and the hole \\
 & \texttt{Drawer Open} & maximise drawer opening \\
\midrule
\multirow{4}{*}{MineDojo}%
 & \texttt{Combat Spider} & combat a spider \\
 & \texttt{Milk Cow} & milk a cow \\
 & \texttt{Shear Sheep} & shear a sheep \\
 & \texttt{Hunt Cow} & hunt a cow \\
\bottomrule
\end{tabular}}
\end{table}

\begin{table}[t]
\centering
\small
\caption{Goal prompts fed to CLIP and MineCLIP encoders.}
\label{tab:clip_prompts}
\resizebox{\linewidth}{!}{%
\begin{tabular}{lll}
\toprule
\textbf{Suite} & \textbf{Task ID} & \textbf{CLIP/MineCLIP prompt} \\
\midrule
Gym            & \texttt{CartPole}              & pole vertically upright on top of the cart. \\[2pt]
\midrule
\multirow{2}{*}{SoftGym}
               & \texttt{Straighten Rope} & The blue rope is straightened. \\
               & \texttt{Pass Water}       & The container, which holds water, is as close to the red circle as possible without causing too many water droplets to spill. \\
\midrule
\multirow{3}{*}{MetaWorld}
               & \texttt{Soccer}                & The soccer ball is in the goal. \\[2pt]
               & \texttt{Sweep Into}            & The green cube is in the hole. \\
               & \texttt{Drawer Open}           & The drawer is opened. \\
\midrule
\multirow{4}{*}{MineDojo}
               & \texttt{Combat Spider}           & combat a spider \\
               & \texttt{Milk Cow}                & milk a cow \\
               & \texttt{Shear Sheep}             & shear a sheep \\
               & \texttt{Hunt Cow}                & hunt a cow \\
\bottomrule
\end{tabular}}
\end{table}

\begin{table}[t]
\centering
\caption{Observation modalities, action spaces, and episode horizons for all evaluated tasks.}
\label{tab:env_specs}
\small
\resizebox{\linewidth}{!}{%
\begin{tabular}{llp{5.0cm}p{4.3cm}c}
\toprule
\textbf{Suite} & \textbf{Task(s)} & \textbf{Observation} & \textbf{Action space} & \textbf{Horizon} \\
\midrule
Gym & CartPole & 4-D state $\langle x,\dot x,\theta,\dot\theta\rangle$ & \texttt{Box}$([0,1])$ – scalar torque & 100 \\
\midrule
\multirow{2}{*}{SoftGym}
  & Straighten Rope & 36-D particle positions & \texttt{picker} – 2 grippers, 6-D $\Delta$ position & 40 (rep.=6) \\
  & Pass Water & 7-D state (container size 3, target pos 1, water-height 1, water‐in/out 2) & \texttt{Box}$(\mathds{R}^{1})$ – $\Delta$ pos along constrained axis & 75 (rep.=8) \\
\midrule
MetaWorld & Soccer, Sweep Into, Drawer Open  & 39-D proprioceptive state (arm 3 + gripper 1 + object 14 + history 18 + goal 3) & \texttt{Box}$(\mathds{R}^{4})$ – $\Delta x,y,z$ + gripper open/close & 500 \\
\midrule
MineDojo & Combat Spider, Milk Cow, Shear Sheep, Hunt Cow & RGB $126{\times}256{\times}3$ (first-person) & \texttt{MultiDiscrete}([12, 3]) – 12 camera/locomotion, 3 mouse magnitudes & 500 \\
\bottomrule
\end{tabular}}
\end{table}

We benchmark our approach on ten goal-conditioned control tasks drawn from four simulators: \textsc{Gym}, \textsc{SoftGym}, \textsc{MetaWorld}, and \textsc{MineDojo}. The tasks are as follows.
\begin{itemize}
    \item \textbf{CartPole}: Balance an upright pole on a moving cart.
    
    \item \textbf{Straighten Rope}: Straighten a rope from a randomly tangled configuration.
    
    \item \textbf{Pass Water}: Move a glass of water to a target location without spilling its contents.
    
    \item \textbf{Open Drawer}: Use a robotic arm to pull open a drawer on a desk.
    
    \item \textbf{Soccer}: Push a soccer ball into a designated goal area using a robotic arm.
    
    \item \textbf{Sweep Into}: Sweep a green cube into a hole on the table using a robotic arm.
    
    \item \textbf{Combat Spider}: Eliminate a spider in combat.  
          The agent is initialized with a diamond sword, a shield, full diamond armor, and a nearby spider.
          
    \item \textbf{Milk Cow}: Find and approach a cow, then collect milk using an empty bucket.  
          The agent is initialized with a bucket, and a cow is spawned nearby.
    
    \item \textbf{Shear Sheep}: Find and approach a sheep, then collect its wool using a shear.  
          The agent starts with a shear and a sheep nearby.
    
    \item \textbf{Hunt Cow}: Find and approach a cow, then hunt it using a sword.  
          The cow will attempt to flee, requiring the agent to chase it down.  
          The agent is initialized with a diamond sword and a cow nearby.
    
\end{itemize}

For each task we hand-craft a concise natural-language \emph{goal prompt} that is fed into the VLM during preference querying (see Appendix \ref{appendix:VLM_prompt}).  
Table \ref{tab:tasks} summarizes the task IDs and the exact goal prompts. 
For image–language models based on \emph{frozen} CLIP or MineCLIP encoders we supply a different
set of goal descriptions phrased as \emph{achievement statements}.
Table~\ref{tab:clip_prompts} lists the exact prompts; 
The prompts adopt a declarative ``goal achieved’’ style
(e.g.\ \emph{``The drawer is opened.''}) that is empirically more compatible with
the image–text contrastive objectives.
Table \ref{tab:env_specs} summarises the observation modality, action space, and episode horizon of each task.

\textbf{Additional notes:}
\begin{itemize}
    \item \textit{Straighten Rope}\,: two end-point pickers are pre-grasped at reset; the 6-D action is the 3-D $\Delta$ position of each picker.
    \item \textit{Pass Water}\,: the glass container moves along a single horizontal axis, resulting in a 1-D action.
    \item \textit{MetaWorld tasks}\,: we initialise the gripper close to the target object and re-align the camera such that the object is centred, producing cleaner images for the VLM.
    \item \textit{MineDojo tasks}\,: observations are limited to first-person RGB frames only (no yaw, pitch, GPS, or voxel data) to better reflect real-world vision. We enable \emph{fast reset} with random teleportation, fix the world seed to~123, and constrain the biome to \textsc{SunflowerPlains}.
\end{itemize}

\section{Details on Baselines}
\label{appendix:detail_baselines}

\subsection{VLM-AR3L (Ours)}
In all experiments, we set the absolute reward weighting to $\alpha = 0.5$, the confidence threshold to $\tau = 0.52$, and the temporal offset to $k = 16$.

\subsection{RL-VLM-F}
Released by \cite{wang2024}, RL-VLM-F employs two image--reward backbones that differ by environment suite:
\begin{itemize}
    \item \textbf{Custom CNN}\,: Table~\ref{tab:cnn_arch}; used for \textsc{Gym} and \textsc{MetaWorld}.
    \item \textbf{ResNet-18}\,: Table~\ref{tab:resnet_arch}; used for \textsc{SoftGym} and \textsc{MineDojo}.
\end{itemize}

\begin{table}[t]
  \centering
  \caption{Custom CNN backbone.}
  \label{tab:cnn_arch}
  \small
  \begin{tabular}{cccc}
    \toprule
    \# & Layer                  & Specification (kernel / stride) & Activation \\ \midrule
    1 & Conv2d $3{\to}16$       & $5{\times}5$ / 3                & ReLU \\
    2 & Conv2d $16{\to}32$      & $3{\times}3$ / 2                & ReLU \\
    3 & Conv2d $32{\to}64$      & $3{\times}3$ / 2                & ReLU \\
    4 & Conv2d $64{\to}128$     & $3{\times}3$ / 2                & ReLU \\
    5 & Flatten                 & —                               & —    \\
    6 & Linear $512{\to}1$      & —                               & Tanh \\
    \bottomrule
  \end{tabular}
\end{table}

\paragraph{ResNet-18 Backbone (SoftGym \& MineDojo).}
A vanilla ResNet-18 (\texttt{BasicBlock}\,$\times[2,2,2,2]$) is adopted for high‐variance visual inputs.
ImageNet initialisation is retained, but the final FC layer is replaced with a 1- or 2-unit head.

\begin{table}[t]
  \centering
  \caption{ResNet-18 backbone used in RL-VLM-F baselines.}
  \label{tab:resnet_arch}
  \small
  \begin{tabular}{cccc}
    \toprule
    Stage & Layer / Block      & Specification                                   & Activation \\ \midrule
    0     & Conv1              & $7{\times}7$, 64, stride 2 + MaxPool            & ReLU \\
    1     & Layer1             & BasicBlock, 64\,$\times$\,2                     & ReLU \\
    2     & Layer2             & BasicBlock, 128\,$\times$\,2                    & ReLU \\
    3     & Layer3             & BasicBlock, 256\,$\times$\,2                    & ReLU \\
    4     & Layer4             & BasicBlock, 512\,$\times$\,2                    & ReLU \\
    5     & AvgPool            & AdaptiveAvgPool2d ($1{\times}1$)                & —    \\
    6     & FC head            & Linear $512{\to}1$               & —    \\
    \bottomrule
  \end{tabular}
\end{table}

\medskip
\noindent\textbf{Training protocol.}  
Except for the backbone choice above, all optimisation hyper-parameters and data-collection schedules follow the settings described in Section~\ref{appendix:siamese}.

\subsection{CLIP}
\label{appendix:baseline_clip}

For image–language matching we use the original \textbf{OpenAI CLIP} ViT-L/14 @ 336 px checkpoint.
During interaction we compute a \emph{static} similarity reward at each step:

\begin{enumerate}
    \item \textbf{Text embedding.}  
          The goal prompt $g$ is tokenised and encoded once; its feature vector is $\ell_{2}$-normalised.
    \item \textbf{Image embedding.}  
          The current RGB observation is passed through the frozen CLIP image encoder, and $\ell_{2}$-normalised.
    \item \textbf{Reward.}  
          The cosine similarity $r_{t} = \langle f_{\text{img}},f_{\text{text}}\rangle$ is used directly as an reward (higher implies closer to the goal).
\end{enumerate}

\subsection{MineCLIP}
\label{appendix:baseline_mineclip}

For long-horizon visual tasks we employ the \textbf{MineCLIP-attn} reward model of \cite{fan2022minedojo}.  
Its inference pipeline consists of three stages:

\begin{enumerate}
    \item \textbf{Frame encoder} – each RGB frame is embedded by a frozen ViT-B/16 backbone.
    \item \textbf{Temporal aggregator} – a 2-layer Transformer ($d_{\text{model}}{=}512$) compresses the $16$-frame sequence into a single video feature $\mathds{v}\!\in\!\mathds{R}^{F}$.
    \item \textbf{Reward head} – an MLP takes $\mathds{v}$ and the tokenised goal prompt $\mathds{g}$ and outputs a two-logit vector $\mathds{z}=[z_{\textsc{pos}},z_{\textsc{neg}}]$; the positive logit is taken as the scalar reward, $r_{t} = z_{\textsc{pos}}$.
\end{enumerate}

\paragraph{Implementation notes.}
\begin{itemize}
    \item A sliding window stores the most recent 16 frames.  
    Only when the window is full do we query MineCLIP; otherwise the reward is set to~0.
    \item Input frames are normalised with the official MineCLIP statistics $(\mu,\sigma)$.
\end{itemize}

\subsection{GT Dense Reward (Oracle)}
\label{appendix:gt_dense}

For ablation‐style oracle comparisons, we define shaped dense rewards (Table~\ref{tab:gt_dense_formulas}) along with their task-specific weighting coefficients (Table~\ref{tab:gt_dense_lambdas}). The meaning of each variable used in these reward definitions is provided in Table~\ref{tab:gt_symbol_meaning}.

\begin{table}[t]
\centering
\caption{Shaped \emph{dense} ground-truth reward for each task
(\(\mathds{1}[\cdot]\) is the indicator).}
\label{tab:gt_dense_formulas}
\footnotesize
\resizebox{\linewidth}{!}{%
\begin{tabular}{lp{11.2cm}}
\toprule
\textbf{Task} & \textbf{Reward $r_t$} \\ \midrule
CartPole        & $-\lvert\theta_t\rvert$ \\[2pt]
Straighten Rope & $\min\,\!\bigl(d_{\text{end}}(t),\,1.1\,L_{\text{rope}}\bigr)$ \\[2pt]
Pass Water      & $-\lambda_{\text{spill}}\dfrac{w_{\text{out}}}{w_{\text{tot}}}
                  -\lambda_{\text{dist}}\lvert x_t-x_{\text{goal}}\rvert$ \\[6pt]
Soccer          & $\displaystyle\max\,\!\Bigl(
                    \lambda_{\text{object}}\mathds{1}[\text{ball grasped}] +
                    \lambda_{\text{in-place}}\mathds{1}[\text{ball in place}],\;
                    \lambda_{\text{success}}\mathds{1}[d_{\text{ball,goal}}<R_{\text{goal}}]\Bigr)$ \\[6pt]
Sweep Into      & $\displaystyle\max\,\!\Bigl(
                    \lambda_{\text{object}}\mathds{1}[\text{cube grasped}] +
                    \lambda_{\text{in-place}}\mathds{1}[\text{cube in hole}],\;
                    \lambda_{\text{success}}\mathds{1}[d_{\text{cube,hole}}<R_{\text{hole}}]\Bigr)$ \\[6pt]
Drawer Open     & $\lambda_{\text{caging}}\!\cdot\!\text{caging} +
                  \lambda_{\text{opening}}\!\cdot\!\text{opening}$ \\[2pt]
Combat Spider   & $\lambda_{\text{attack}}\mathds{1}[\text{valid attack}] +
                  \lambda_{\text{success}}\mathds{1}[\text{spider eliminated}]$ \\[2pt]
Milk Cow        & $\lambda_{\text{nav}}\max(d_{\min,t-1}-d_{\min,t},0) +
                  \lambda_{\text{success}}\mathds{1}[\text{milk collected}]$ \\[2pt]
Shear Sheep     & $\lambda_{\text{nav}}\max(d_{\min,t-1}-d_{\min,t},0) +
                  \lambda_{\text{success}}\mathds{1}[\text{sheep sheared}]$ \\[2pt]
Hunt Cow        & $\lambda_{\text{attack}}\mathds{1}[\text{valid attack}] +
                  \lambda_{\text{nav}}\max(d_{\min,t-1}-d_{\min,t},0) +
                  \lambda_{\text{success}}\mathds{1}[\text{cow eliminated}]$ \\
\bottomrule
\end{tabular}}
\end{table}

\begin{table}[t]
\centering
\caption{Non-zero coefficients \(\lambda\) for the dense rewards.
Tasks not listed (CartPole, Straighten Rope) have no tunable \(\lambda\).}
\label{tab:gt_dense_lambdas}
\small
\begin{tabular}{ll}
\toprule
\textbf{Task} & \textbf{Coefficients} \\ \midrule
Pass Water      & $\lambda_{\text{spill}}{=}10,\;\lambda_{\text{dist}}{=}1$ \\[2pt]
Soccer          & $\lambda_{\text{object}}{=}3,\;\lambda_{\text{in-place}}{=}6.5,\;\lambda_{\text{success}}{=}10$ \\[2pt]
Sweep Into      & $\lambda_{\text{object}}{=}2,\;\lambda_{\text{in-place}}{=}6,\;\lambda_{\text{success}}{=}10$ \\[2pt]
Drawer Open     & $\lambda_{\text{caging}}{=}5,\;\lambda_{\text{opening}}{=}5$ \\[2pt]
Combat Spider   & $\lambda_{\text{attack}}{=}1,\;\lambda_{\text{success}}{=}10$ \\[2pt]
Milk Cow        & $\lambda_{\text{nav}}{=}0.1,\;\lambda_{\text{success}}{=}10$ \\[2pt]
Shear Sheep     & $\lambda_{\text{nav}}{=}0.1,\;\lambda_{\text{success}}{=}10$ \\[2pt]
Hunt Cow        & $\lambda_{\text{attack}}{=}1,\;\lambda_{\text{nav}}{=}0.1,\;\lambda_{\text{success}}{=}10$ \\
\bottomrule
\end{tabular}
\end{table}

\begin{table}[t]
\centering
\caption{Meaning of the variables used in dense-reward formulas.}
\label{tab:gt_symbol_meaning}
\resizebox{0.9\linewidth}{!}{%
\begin{tabular}{lp{9.0cm}}
\toprule
\textbf{Symbol} & \textbf{Description} \\ \midrule
$\theta_t$                    & Pole angle at time~$t$ (CartPole). Zero means perfectly upright. \\[2pt]

$d_{\text{end}}(t)$           & Distance between the two rope end-points at time~$t$ (Straighten Rope). \\[2pt]
$L_{\text{rope}}$             & Nominal (fully straight) rope length (Straighten Rope). \\[2pt]

$w_{\text{out}}$              & Number of water particles outside the glass (Pass Water). \\[2pt]
$w_{\text{tot}}$              & Total number of water particles (Pass Water). \\[2pt]
$x_t$                         & Current $x$-position of the glass (Pass Water). \\[2pt]
$x_{\text{goal}}$             & Target $x$-position for the glass (Pass Water). \\[4pt]

$d_{\text{ball,goal}}$        & Euclidean distance from soccer ball to goal centre (Soccer). \\[2pt]
$R_{\text{goal}}$             & Acceptable radius around the goal centre (Soccer). \\[4pt]

$d_{\text{cube,hole}}$        & Distance from cube centre to hole centre (Sweep Into). \\[2pt]
$R_{\text{hole}}$             & Acceptable radius around the hole centre (Sweep Into). \\[4pt]

\textit{caging}               & Tolerance-shaped reward encouraging the gripper TCP to \emph{cage} the drawer handle  
                               (Drawer Open).  Formally  
                               $\text{caging}=f\!\bigl(\lVert\mathbf{x}_{\text{grip}}-\mathbf{x}_{\text{handle}}\rVert\bigr)$;  
                               see \texttt{reward\_utils.tolerance}. \\[4pt]

\textit{opening}              & Tolerance-shaped reward for the handle’s displacement toward the fully-open target  
                               (Drawer Open). \\[4pt]

$d_{\min,t}$                  & Shortest agent-to-animal distance at time~$t$  
                               (Milk Cow, Shear Sheep, Hunt Cow). \\[4pt]

$\mathds{1}[\cdot]$           & Indicator function, equals~1 if the event inside the brackets is true, 0 otherwise. \\
\bottomrule
\end{tabular}}
\end{table}

\subsection{GT Sparse Reward (Oracle)}
\label{appendix:gt_sparse}

Sparse rewards remove shaped terms and give a single bonus  
\(r_t = \lambda_{\text{success}}\mathds{1}[\text{task success}]\)
(Table~\ref{tab:gt_sparse_success}).

\begin{table}[!t]
\centering
\caption{Success criteria and sparse-reward scale \(\lambda_{\text{success}}\).}
\label{tab:gt_sparse_success}
\small
\begin{tabular}{lll}
\toprule
\textbf{Task} & \textbf{Success condition} & \(\lambda_{\text{success}}\) \\ \midrule
CartPole        & $\lvert\theta_t\rvert < 5^{\circ}$ & 1 \\[2pt]
Straighten Rope & $d_{\text{end}} > 0.95\,L_{\text{rope}}$ & 1 \\[2pt]
Pass Water      & Dense reward $>-0.36$ & 1 \\[2pt]
Soccer          & ball within $R_{\text{target}}$ of goal & 1 \\[2pt]
Sweep Into      & cube within $R_{\text{target}}$ of hole & 1 \\[2pt]
Drawer Open     & drawer opened beyond $R_{\text{target}}$ & 1 \\[2pt]
Combat Spider   & spider eliminated & 10 \\[2pt]
Milk Cow        & milk collected & 10 \\[2pt]
Shear Sheep     & wool collected & 10 \\[2pt]
Hunt Cow        & cow eliminated & 10 \\
\bottomrule
\end{tabular}
\end{table}

\section{Additional Experimental Results}

\subsection{Ablation Hyper-parameters}
\label{appendix:ablation_hyper}
\begin{figure}[!t]
  \centering
  \includegraphics[width=0.8\linewidth]{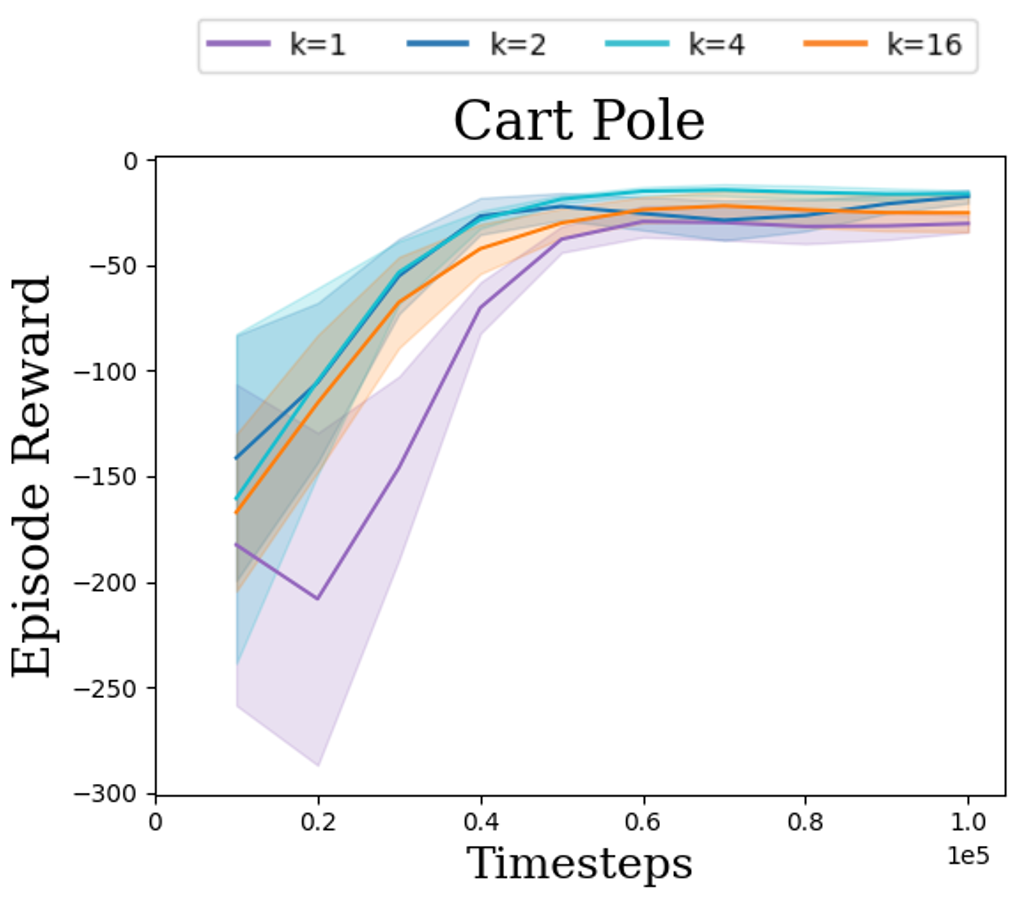}
  \caption{\textbf{Ablation studies of temporal offset $k$.}}
  \label{fig:ablation_k}
\end{figure}

\begin{figure}[!t]
  \centering
  \includegraphics[width=1\linewidth]{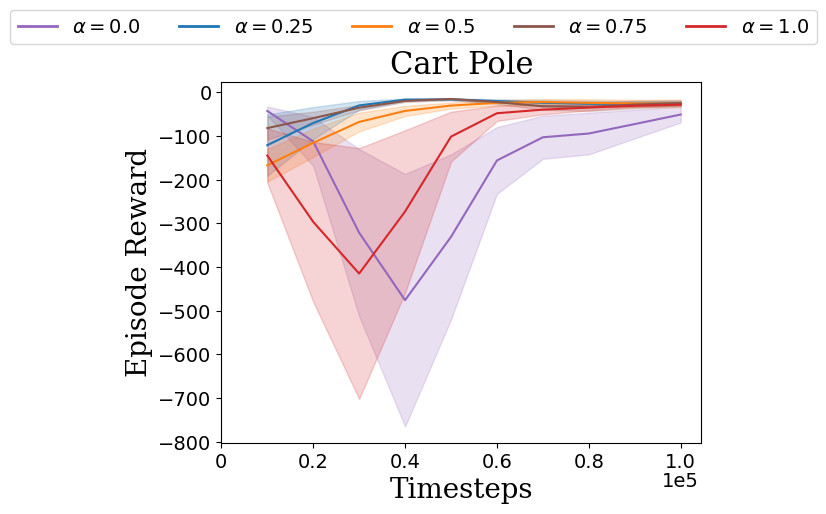}
  \caption{\textbf{Ablation studies of weighting coefficient $alpha$.}}
  \label{fig:ablation_alpha}
\end{figure}

We conduct ablation studies on the temporal offset $k$ and the weighting coefficient $\alpha$. As shown in Fig.~\ref{fig:ablation_k}, varying $k$ does not result in significant performance differences, indicating that our method is relatively insensitive to the choice of temporal offset. Fig.~\ref{fig:ablation_alpha} investigates the weighting coefficient $\alpha$ for the absolute reward signals. The method performs consistently well over a broad range of $\alpha$, while performance deteriorates only when extremely small or large values are used.

\clearpage

\begin{figure*}[!t]
  \centering
  \includegraphics[width=0.75\linewidth]{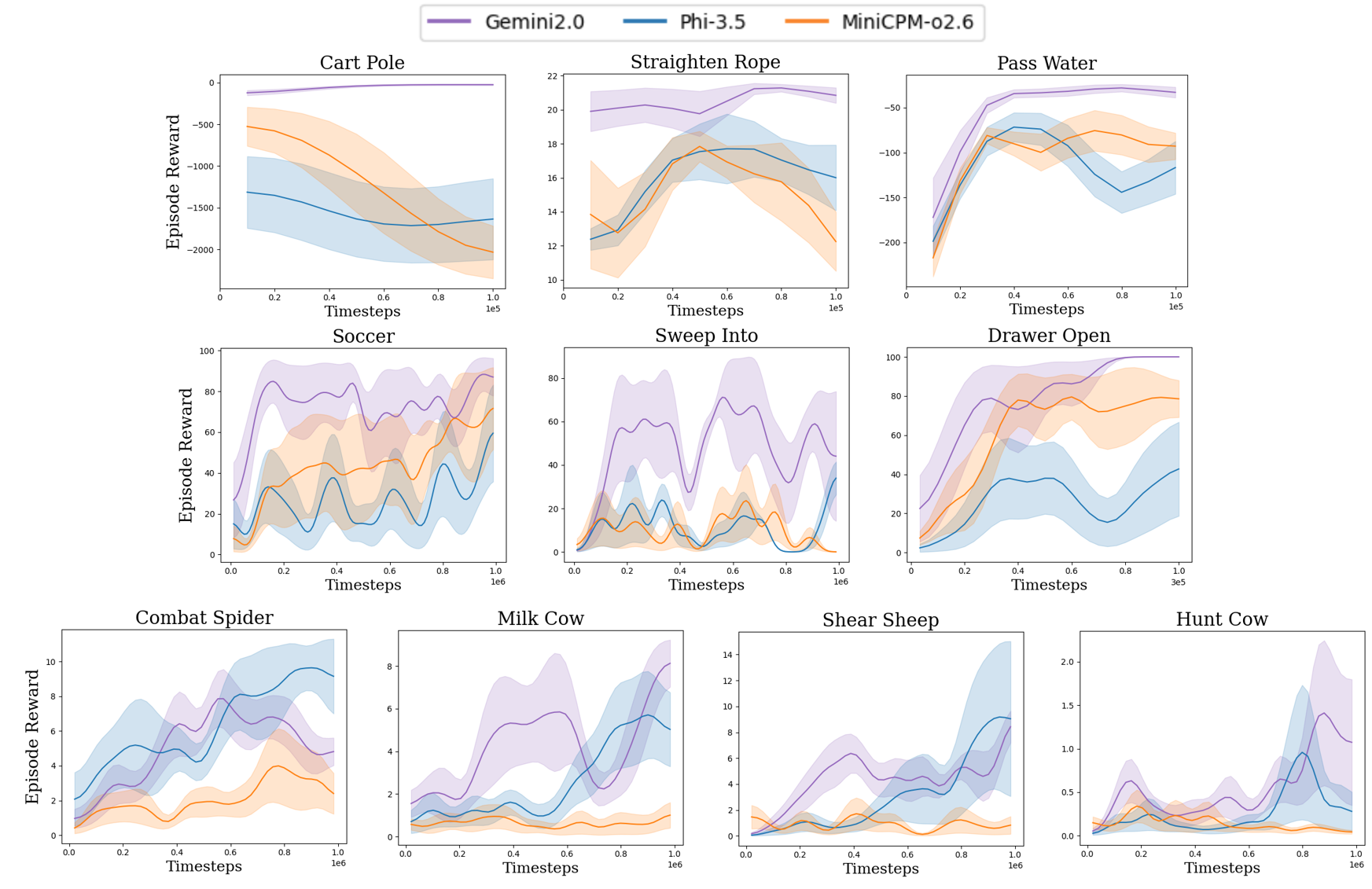}
  \caption{Ablation results with different vision-language models.}
  \label{fig:ablation_VLM}
\end{figure*}

\subsection{Ablation VLM}
\label{appendix:ablation_vlm}
See Figure~\ref{fig:ablation_VLM} and Table~\ref{tab:ablation_vlm}.
Our results show that Phi-3.5 exhibits strong performance in interpreting Minecraft-specific visual states. MiniCPM-o 2.6, on the other hand, performs competitively on more structured tasks such as Soccer and Drawer Open.

\subsection{Ablation Absolute and Relative rewards}
\label{appendix:ablation_combine}
\begin{figure*}[!t]
  \centering
  \includegraphics[width=0.75\linewidth]{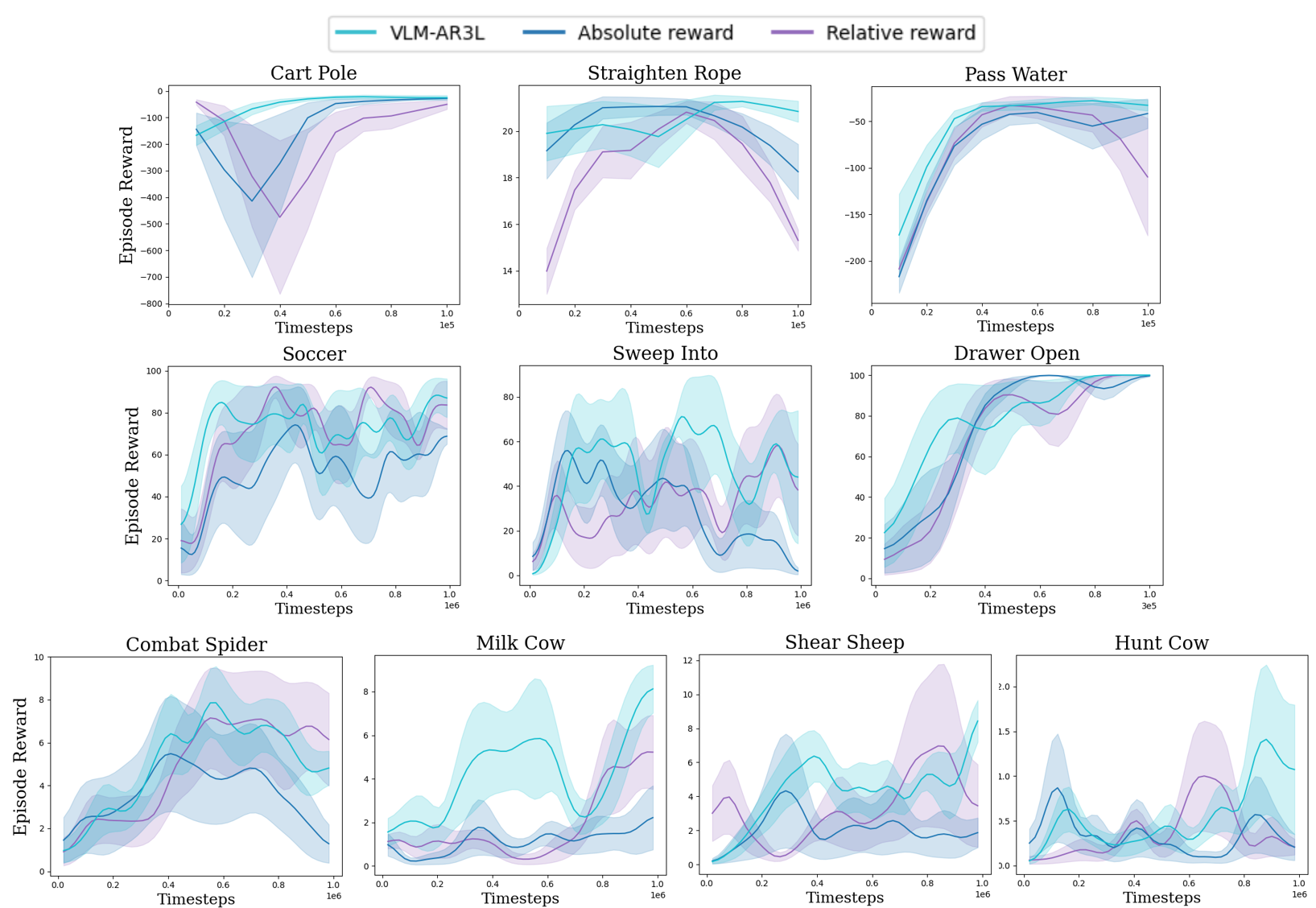}
  \caption{Ablation studies of absolute and relative rewards.}
  \label{fig:ablation_combine}
\end{figure*}

See Figure~\ref{fig:ablation_combine}. We compare three variants: absolute-only ($\alpha{=}1$), relative-only ($\alpha{=}0$), and the dual reward (default $\alpha{=}0.5$). Absolute rewards tend to work well in tasks with a clear global progress measure, while relative rewards provide more reliable guidance in long-horizon or visually ambiguous tasks. Combining them yields the best or near-best performance in most tasks, supporting the complementarity discussed in Section~\ref{sec:method}.

\section{Broader Impact}
\label{appendix:broader_impact}

Our work leverages vision-language models (VLMs) to provide reward signals for reinforcement learning without requiring extensive manual reward engineering. This approach has several positive societal implications, including:

\begin{itemize}
    \item \textbf{Reducing Development Costs:} By using pre-trained VLMs to generate rewards, we reduce the need for domain-specific reward design, potentially lowering the cost and effort required to train RL agents in complex or open-ended environments.
    \item \textbf{Scalability:} The method can be applied to various domains without extensive task-specific tuning, facilitating broader adoption of RL in areas like robotics, game environments, and simulation-based training.
\end{itemize}

However, there are also potential negative implications:

\begin{itemize}
    \item \textbf{Bias Propagation:} If the VLMs used for reward labeling exhibit biases, these biases could be propagated to the RL agents, potentially leading to biased or harmful behaviors.
    \item \textbf{Unintended Applications:} The method could be exploited to train agents for undesirable purposes, such as manipulating environments to achieve harmful objectives or reinforcing negative behaviors.
\end{itemize}

To mitigate these risks, we recommend implementing bias detection mechanisms in VLM outputs and restricting the application of the trained RL agents to controlled and monitored environments. Additionally, transparency in VLM training procedures can further minimize potential harms.
\begin{table*}[!t]
\centering
\caption{Success rate (mean $\pm$ standard error) of policy using difference VLM. Results are averaged over 3 random seeds with 5 evaluation episodes per best checkpoint.}
\label{tab:ablation_vlm}
{
\small
\resizebox{\linewidth}{!}{
\begin{tabular}{l|c|cc|ccc|cccc}
\toprule
\textbf{Methods} & \textbf{CartPole} & \textbf{Straighten} & \textbf{Pass}
& \textbf{Soccer} & \textbf{Sweep} & \textbf{Drawer} & \textbf{Spider} & \textbf{Milk} & \textbf{Sheep} & \textbf{Hunt} \\
& & \textbf{Rope} & \textbf{Water}
&  & \textbf{Into} & \textbf{Open} & \textbf{Combat} & \textbf{Cow} & \textbf{Shear} & \textbf{Cow} \\
\midrule
Gemini2.0 & \textbf{1.00 $\pm$ 0.00}& \textbf{1.00 $\pm$ 0.00}& \textbf{1.00 $\pm$ 0.00}& \textbf{1.00 $\pm$ 0.00}& \textbf{1.00 $\pm$ 0.00}& \textbf{1.00 $\pm$ 0.00}& \textbf{0.85 $\pm$ 0.05}& \textbf{0.95 $\pm$ 0.00}& \textbf{0.70 $\pm$ 0.10}& \textbf{0.52  $\pm$ 0.02}\\

Phi3.5    & 0.67 $\pm$ 0.47& 1.00 $\pm$ 0.00& 0.33 $\pm$ 0.47& 1.00 $\pm$ 0.00& 1.00 $\pm$ 0.00& 1.00 $\pm$ 0.00& 0.83 $\pm$ 0.13& 0.81 $\pm$ 0.18& 0.53 $\pm$ 0.34& 0.37 $\pm$ 0.29\\

MiniCPM-o2.6 & 1.00 $\pm$ 0.00& 1.00 $\pm$ 0.00& 0.67 $\pm$ 0.47& 1.00 $\pm$ 0.00& 1.00 $\pm$ 0.00& 1.00 $\pm$ 0.00& 0.52 $\pm$ 0.23& 0.43 $\pm$ 0.09& 0.33 $\pm$ 0.19& 0.19 $\pm$ 0.09\\

\bottomrule
\end{tabular}
}}
\end{table*}